\documentclass{article}

\usepackage{microtype}
\usepackage{graphicx}
\usepackage{subfigure}
\usepackage{booktabs} %
\usepackage{multirow}

\usepackage{hyperref}

\usepackage[accepted]{icml2021}

\icmltitlerunning{Technical Challenges for Training Fair Neural Networks}

\usepackage{misc_symbols}
\usepackage{times}
\usepackage{epsfig}
\usepackage{graphicx}
\usepackage{amsmath}
\usepackage{amssymb}
\usepackage{arydshln}
\usepackage{xspace}
\usepackage{pifont}
\usepackage{hhline}
\usepackage{amsthm}
\usepackage[dvipsnames, table]{xcolor}
\usepackage{accents}
\usepackage{color, colortbl}
\usepackage{multirow}

\begin{document}

\twocolumn[
\icmltitle{Technical Challenges for Training Fair Neural Networks}

\icmlsetsymbol{equal}{*}

\begin{icmlauthorlist}
\icmlauthor{Valeriia Cherepanova}{equal,umd}
\icmlauthor{Vedant Nanda}{equal,umd,mpi}
\icmlauthor{Micah Goldblum}{umd}
\icmlauthor{John P. Dickerson}{umd}
\icmlauthor{Tom Goldstein}{umd}
\end{icmlauthorlist}

\icmlaffiliation{umd}{University of Maryland, College Park, USA}
\icmlaffiliation{mpi}{Max Planck Institute for Software Systems, Germany}

\icmlcorrespondingauthor{Valeriia Cherepanova}{vcherepa@umd.edu}
\icmlcorrespondingauthor{Vedant Nanda}{vedant@cs.umd.edu}

\icmlkeywords{Machine Learning, ICML}

\vskip 0.3in
]

\printAffiliationsAndNotice{\icmlEqualContribution} %

\begin{abstract}

As machine learning algorithms have been widely deployed across applications, many concerns have been raised over the fairness of their predictions, especially in high stakes settings (such as facial recognition and  medical imaging).  To respond to these concerns, the community has proposed and formalized various notions of fairness as well as methods for rectifying unfair behavior.  While fairness constraints have been studied extensively for classical models, the effectiveness of methods for imposing fairness on deep neural networks is unclear.  In this paper, we observe that these large models overfit to fairness objectives, and produce a range of unintended and undesirable consequences. We conduct our experiments on both facial recognition and automated medical diagnosis datasets using state-of-the-art architectures.

\end{abstract}

\vspace{-7mm}

\begin{figure*}[t]
\centering
\includegraphics[width=0.84\textwidth]{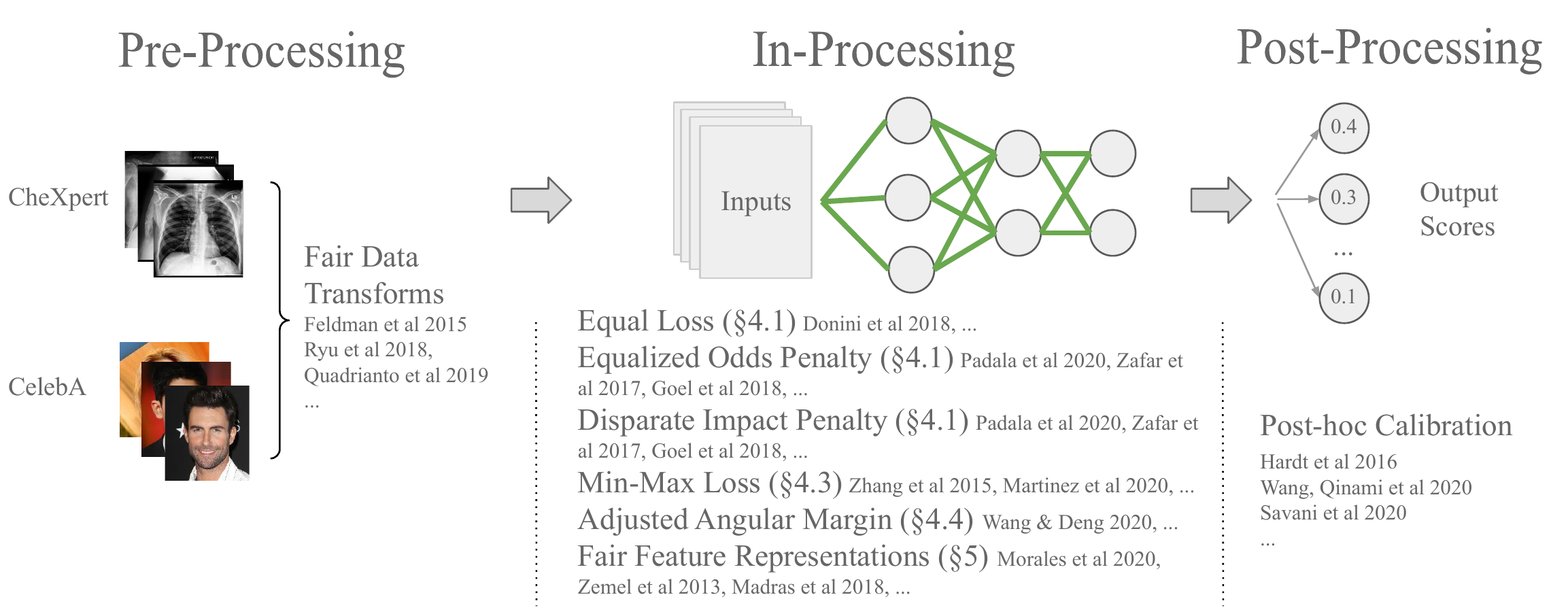}
\caption{\textbf{[Brief Overview of Fairness in ML]} 
For the scope of this paper, we consider only in-processing techniques and apply them to deep neural networks. We show that that overparametrized nature of neural networks is one reason why current techniques fail.}
\label{fig1}
\vskip -0.1in
\end{figure*}

\section{Introduction}
\label{sec:intro}

Machine learning systems are increasingly deployed in settings with major economic and social impacts.  In such situations, differences in model behaviors between different social groups may result in disparities in how these groups are treated~\cite{barocas2016big,sainyam2017fairness}.
For this reason, it is crucial to understand the bias of machine learning systems, and to develop tools that prevent algorithmic discrimination against protected groups.  

Much effort has been devoted to understanding and correcting biases in classical machine learning models (\eg~SVMs, Logistic Regression \etc).  Overfitting is not a pernicious issue for classical models, and so fairness constraints that are imposed at train time often generalize to (unseen) test-time data. For overparameterized neural networks -- as is often the case with modern deep neural networks~\cite{zhang2017understanding} -- our tools for understanding and controlling model bias are far less effective, in large part because of the difficulties created by overfitting.  At train time, neural networks interpolate the data and achieve perfect accuracy on all sub-groups, thus making it impossible to obtain meaningful measures of bias on training data. When constraints are imposed using a sequestered dataset, the network may still overfit to constraints. Furthermore, the extremely fluid decision boundaries of neural networks open the possibility for complex forms of {\em fairness gerrymandering}~\cite{kearns2018preventing}, in which the decision boundaries of the model are moved to achieve a fairness constraint, while at the same time creating unintended consequences for important sub-groups that were not explicitly considered at train time.  Since deep neural networks perform so much better than their linear counterparts on a wide range of tasks, it is important that we better understand the fairness properties of these complex systems.

This paper investigates whether currently available methods for training fair models are capable of controlling bias in Deep Neural Networks (DNNs).  We find that train-time fairness interventions -- including those that have been thoroughly tested for classical ML models -- are not effective for DNNs.
We further show that when these fairness interventions do seem to work, they often result in the undesired phenomenon of fairness gerrymandering.
Our contributions are as follows:

\begin{itemize}
\item We empirically test a range of existing methods for imposing fairness using constraints and penalties during training of DNNs. While methods of this type have been widely used and rigorously studied in the under-fitting regime (\ie{} SVMs and linear models), we show that they fail in the overparameterized regime. 
\item We find that in some cases, fairness surrogates that work well for classical models do not work well for DNNs. In particular, equality of losses does not necessarily translate to equality of metrics used to evaluate performance (\eg{} area under the curve). 
\item  We also observe that specialized constraints designed for facial recognition often appear to work on training data, but fail on holdout classes. Facial recognition systems present a unique case because they operate differently during inference than during training.
\item  We consider adversarial methods for learning ``fair features.''  In addition to discussing theoretical problems with these approaches, we observe that these methods are not effective at achieving fair outcomes in practice.
\item We observe that fairness gerrymandering can be particularly problematic for DNNs because of their highly flexible decision boundaries. That is, parity along one sensitive attribute (\eg{} sex) comes at the cost of increased disparity along another sensitive attribute (\eg{} age).

\end{itemize}

\xhdr{Related Work} There exist well documented cases of unfairness in key ML applications such as targeted ads~\cite{speicher2018potential,ali2019discrimination,ribeiro2019microtagging}, personalized recommendation systems~\cite{beiga2018equity, singh2018fairness}, credit scoring~\cite{khandani2010consumer}, recidivism prediction~\cite{chouldechova2017Fair}, hiring~\cite{schumann2020hiring}, medical diagnosis~\cite{larrazabal2020gender, seyyedkalantari2020chexclusion}, facial recognition~\cite{buolamwini2018gendershades,grother2010report,ngan2015face}, and others. This has resulted in a range of interdisciplinary work on understanding and mitigating bias in automated decision making systems~\cite{binns2017fairness,Leben2020Normative,hashimoto2018fairness,martinez2020minimax,nanda2021fairness,heidari2019on}.

Existing work on mitigating algorithmic unfairness can be broadly put into three categories: pre-processing, in-processing and post-processing (see Figure~\ref{fig1}). Works on pre-processing mostly focus on removing sensitive information from the data and building diverse and balanced datasets~\cite{Feldman2015Certifying,ryu2018inclusivefacenet,quadrianto2019discovering,wang2020mitigating}. In-processing aims to change the training routine, often via imposing constraints~\cite{zafar2017aistats,Zafar2017www,zafar2019jmlr,donini2018empirical,goel2018non,Padala2020achieving,agarwal2018reductions,wang2020mitigating} or by changing the optimization routine~\cite{martinez2020minimax,diana2020convergent,lahoti2020fairness}. Another in-processing strategy is to learn fair intermediate representations independent of sensitive attributes which lead to fairness on any downstream task~\cite{dwork2012fairness,zemel13learning,edwards2016censoring,madras2018learning,beutel2017data,wang2019balanced}. Post-processing techniques aim to change the inference mechanism to ensure fair outcomes~\cite{hardt2016equality,wang2020fairness,savani2020posthoc}. In this paper, we limit our scope to understanding how in-processing techniques work for DNNs.

Prior work has also suggested that algorithmic solutions for fairness are often hard to comprehend~\cite{Saha20:Measuring} or put into practice~\cite{beutel2019putting}. Additionally, industry practitioners believe that fairness issues in real-world ML systems could be more effectively tackled by systematically changing the broader system design, rather than solely focusing on algorithmic ``debiasing''~\cite{holstein2019improving,Madaio20:Co-designing}. Our work aims to show the fragility of algorithmic fairness interventions for deep models, thus extending the scholarship on challenges for fairness in ML.

\section{Experimental Setup}
\label{sec:setup}
In this work, we investigate how different in-processing methods for mitigating algorithmic unfairness work on two different problems: facial recognition and medical image classification. We choose these two particular domains to illustrate broadly applicable pitfalls when applying train-time fairness interventions in deep learning. Moreover, while both facial recognition and medical image classification have seen unprecedented performance gains due to advances in deep learning, there have been well documented cases of bias and unfairness in both of these domains~\cite{grother2010report,buolamwini2018gendershades,seyyedkalantari2020chexclusion,larrazabal2020gender}. Additionally, previous works have also outlined ethical and epistemic issues with facial recognition and algorithmic solutions to fairness in healthcare~\cite{mccradden2020when,raji2020saving,andrejevic2020facial, cherepanova2021lowkey}. 

In this section, we describe our experimental setup which we use throughout the paper. 

\subsection{Face Recognition} 
\label{sec:setup_fr}
\par State-of-the-art facial recognition (FR) systems contain deep neural networks that extract facial features from {\bf probe images} (new photos whose subject is identified by FR) and compare the resulting features to those corresponding to {\bf gallery images} (references with known identities). Probe images are matched to the gallery image whose features lie closest with respect to the Euclidean metric. In order to distinguish a vast number of identities, these large systems are trained using special classification heads, such as ArcFace and CosFace, designed to increase angular margins between identities, and these heads are removed for inference ~\cite{deng2019arcface, wang2018cosface, liu2017sphereface}.

\par We train facial recognition models with ResNet-18 and ResNet-152 backbones using the popular CosFace head. All of our models are trained as classifiers using focal loss~\cite{lin2017focal}. %
We use the Celeb-A dataset, which contains labels for age, gender, and non-sensitive attributes such as ``Eyeglasses'' or ``Wearing hat''. We split the Celeb-A dataset into train and test sets with non-overlapping identities. Therefore, at test time, the FR model is evaluated on people whose images it has not seen during training.  We have 9,177 training identities (4058 male and 5119 female) with over 185,000 images and 9,177 testing images from 1000 identities (487 male and 513 female). Finally, we split a validation set from the training data consisting of 3 images from each identity with more than 6 images. 

\par We measure the performance of FR systems with two methods. On training data we can use the classification head from training to report multi-label classification accuracy, while for validation and test sets we rip off the classification head and report rank-1 nearest neighbor (in feature space) accuracy as is mainstream in the facial recognition literature. When the nearest-neighbor match for an image is the correct identity, we call this a \emph{true positive}. We find that standard facial recognition models exhibit lower testing accuracy for females than for males, see Table~\ref{tab:adjusted_margins_table}. This accuracy gap does not result from unbalanced data; in fact female identities have more gallery images on average than male identities and 56\% of identities in the data are female.

\par 

\subsection{Medical Image Classification}
We use CheXpert~\cite{irvin2019chexpert}, a widely used and publicly available benchmark dataset for chest X-ray classification. This dataset consists of 224,316 chest radiographs annotated with a binary label indicating the presence of a given pathology. We consider the following 5 pathologies: Cardiomegaly (CA), Edema (ED), Consolidation (CO), Atelectasis (AT) and Pleural Effusion (PE). This yields a \textit{multi-label} classification task, predicting which of the 5 pathologies are present given a chest x-ray image. We train models using weighted binary cross entropy loss, and we report performance via area under the curve (AUC) for each of the 5 tasks.

Our experiments only use images for which sex and age labels are available, yielding a total of 223,413 images. We randomly split this data in a 80:20 ratio to form the training and validation sets respectively. The dataset also provides a test set with $234$ images labelled by radiologists. The training set is primarily composed of males (60\%), while validation and test sets are more balanced (55\% males).

We use the highest ranked model on the CheXpert leaderboard\footnote{https://stanfordmlgroup.github.io/competitions/chexpert/} with a publicly available implementation.\footnote{https://github.com/jfhealthcare/Chexpert} We fine-tune a Densenet121~\cite{huang2017densely} pre-trained on ImageNet~\cite{russakovsky2015imagenet} for this task. 
Additional details about the model, data preprocessing, optimizer, and hyperparameters can be found in Appendix~\ref{sec:appendix_setup}.

\section{Fairness Notions}
\label{sec:notions}

We consider the traditional fair machine learning setup consisting of a training dataset $\Dcal = \{(x_{i}, a_{i}, y_{i})\}_{i=1}^{N}$, where $x_i$ are drawn independently from the input distribution $\Xcal$, $a_i \in \Acal$ are sensitive features (such as race, gender \etc) and $y_{i} \in \Ycal$ are true labels. For simplicity's sake, assume that $\Acal = \{0, 1\}$. $\Ycal$ is binary in the medical imaging task and multi-class for facial recognition. We wish to train a model $f_{\theta} : \Xcal \rightarrow \RR$ which can predict an outcome $\hat{y_i}$ for a given $x_i$.
Standard training procedures outside of the fair training regime
minimize the average loss over training samples, $\hat{L}(f_{\theta})$. We refer to this as the \textit{baseline} training scheme in our experiments. We refer to average loss on data points where $a_i = 1$ as $\hat{L}^{a+}(f_{\theta})$ and points where $a_i = 0$ as $\hat{L}^{a-}(f_{\theta})$. Below, we recall fairness notions which we use in this work and also describe how we operationalize them.

{\bf Accuracy Equality} requires the classification system to have equal misclassification rates across sensitive groups~\cite{zafar2017aistats,Zafar2017www,zafar2019jmlr}. 
\begin{equation}
\mathbb{P} (\hat{y} \ne y|a = 0) \approx \mathbb{P} (\hat{y} \ne y|a = 1)
\end{equation}
Because accuracy is a discontinuous function of the model parameters, we use equal loss as a surrogate, and solve:
    \begin{equation}
    \label{eq:eq_loss}
        \min_{\theta} \bigg[ \hat{L}(f_{\theta}) + \alpha  | \hat{L}^{a+}(f_{\theta}) - \hat{L}^{a-}(f_{\theta}) | \bigg]
    \end{equation}

{\bf Equalized Odds} aims to equalize the true positive and false positive rates for a classifier (sometimes also referred to as \textit{disparate mistreatment})~\cite{hardt2016equality}.
\begin{equation}
\mathbb{P} (\hat{y} = 1 | a = 1, y = y) \approx \mathbb{P} (\hat{y} = 1 | a = 0, y = y)
\end{equation}
 The Equalized Odds Penalty~\cite{Padala2020achieving} aims to approximate equalized odds using logits as measures of probability. Thus, we minimize the following objective:
    \begin{equation}
    \label{eq:eq_odds}
        \min_{\theta} \bigg[ \hat{L}(f_{\theta}) + \alpha (fpr + fnr) \bigg],
    \end{equation}
    where
    \begin{equation*}
        fpr = \bigg| \frac{\sum_i p_i(1-y_i)a_i}{\sum_i a_i} - \frac{\sum_i p_i(1-y_i)(1-a_i)}{\sum_i (1-a_i)} \bigg|
    \end{equation*}
    \begin{equation*}
        fnr = \bigg| \frac{\sum_i (1-p_i)y_ia_i}{\sum_i a_i} - \frac{\sum_i (1-p_i)y_i(1-a_i)}{\sum_i (1-a_i)} \bigg|.
    \end{equation*}
Here, $p_i$ denotes a softmax output (binary prediction task).

{\bf Disparate Impact} is a widely adopted notion that requires any decision making process' outcomes to be independent of membership in a sensitive group~\cite{calders2009building,barocas2016big,chouldechova2017Fair,Feldman2015Certifying}:
\begin{equation}
\mathbb{P} (\hat{y}=1 | a = 1) \approx \mathbb{P} (\hat{y} = 1 | a = 0)
\end{equation}
The Disparate Impact Penalty~\cite{Padala2020achieving} aims to approximate disparate impact through the objective,
    \begin{equation}
    \label{eq:di}
        \min_{\theta} \bigg[ \hat{L}(f_{\theta}) + \alpha di \bigg],
    \end{equation}
    where
    \begin{equation*}
        di = -\min\left( \frac{\frac{\sum_i a_ip_i}{\sum_i a_i}}{\frac{\sum_i (1 - a_i)p_i}{\sum_i (1 - a_i)}}, \frac{\frac{\sum_i (1 - a_i)p_i}{\sum_i (1 - a_i)}}{\frac{\sum_i a_ip_i}{\sum_i a_i}}  \right).
    \end{equation*}

{\bf Max-Min Fairness} focuses on maximizing the performance for the most discriminated against group, i.e. the group with lowest utility~\cite{rawls1999theory,zhang2014fairness,hashimoto2018fairness,mohri2019agnostic,martinez2020minimax,diana2020convergent,lahoti2020fairness}. 
\begin{equation}
\max \min_{a\in \Acal} \mathbb{P} (\hat{y}=y |a)
\end{equation}
To optimize models for Max-Min fairness, we minimize the loss for the sensitive group with maximum loss at the current iteration. That is, we perform the following optimization at each iteration: 
\begin{equation*}
    \min_{\theta}~ \max ~\{\hat{L}^{a+}(f_{\theta}), \hat{L}^{a-}(f_{\theta})\}
\end{equation*}
 Since both equality of opportunity and disparate impact notions assume existence of a beneficial outcome, they are most useful for binary classification tasks, and we only use them in the medical image classification task.

\begin{table*}[t]
\caption{\textbf{[CheXpert]} AUC shown for 2 tasks, Consolidation (CO) and Pleural Effusion (PE), along with the average over all 5 tasks. 
More detailed results are in Appendix~\ref{sec:appendix_results}. 
We observe that due to their tendency to overfit the training data, adding regularizers to DNNs on training data can actually hurt disparity on the test set. 
One notable exception is the equal loss regularizer, which achieves better parity on the test set. 
However, as we discuss in Section~\ref{sec:gerrymandering}, this model shows \textit{fairness gerrymandering}. 
Additionally, we observe that optimizing for the regularizer on a holdout validation set or using a min-max loss does not lead to less disparity on the test set.}
\label{tab:chexpert_reduced_results}
\vskip 0.15in
\begin{center}
\begin{small}
\begin{sc}
\begin{tabular}{c|c|ccc|ccc|ccc}
\toprule
\multirow{2}{*}{Scheme}                                                                 & \multirow{2}{*}{Task} & \multicolumn{3}{c|}{Train} & \multicolumn{3}{c|}{Validation} & \multicolumn{3}{c}{Test} \\
                                                                                        &                          & Male   & Female   & Gap    & Male     & Female     & Gap     & Male   & Female   & Gap  \\ 
\midrule
\midrule

\multirow{3}{*}{\begin{tabular}[c]{@{}c@{}}Baseline\end{tabular}} 
& CO & 0.996 & 0.996 & 0.000 & 0.681 & 0.687 & 0.006 & 0.896 & 0.808 & 0.088 \\ 
& PE & 0.936 & 0.942 & 0.006 & 0.841 & 0.854 & 0.013 & 0.895 & 0.925 & 0.030 \\ 
& \textbf{Avg} & \textbf{0.950} & \textbf{0.953} & \textbf{0.002} & \textbf{0.753} & \textbf{0.753} & \textbf{0.007} & \textbf{0.816} & \textbf{0.781} & \textbf{0.047} \\ 

\hline

\multirow{3}{*}{\begin{tabular}[c]{@{}c@{}}Eq. Loss \\Train \end{tabular}} 
& CO & 0.983 & 0.981 & 0.003 & 0.699 & 0.698 & 0.000 & 0.835 & 0.819 & 0.016 \\ 
& PE & 0.900 & 0.906 & 0.006 & 0.850 & 0.864 & 0.015 & 0.886 & 0.946 & 0.059 \\
& \textbf{Avg} & \textbf{0.907} & \textbf{0.907} & \textbf{0.003} & \textbf{0.769} & \textbf{0.770} & \textbf{0.005} & \textbf{0.847} & \textbf{0.843} & \textbf{0.028} \\ 

\hline

\multirow{3}{*}{\begin{tabular}[c]{@{}c@{}}Eq. Odds\\Penalty \end{tabular}}  
& CO & 0.998 & 0.998 & 0.000 & 0.659 & 0.664 & 0.004 & 0.781 & 0.665 & 0.116 \\ 
& PE & 0.928 & 0.934 & 0.005 & 0.823 & 0.836 & 0.013 & 0.769 & 0.863 & 0.094 \\ 
& \textbf{Avg} & \textbf{0.954} & \textbf{0.956} & \textbf{0.002} & \textbf{0.739} & \textbf{0.742} & \textbf{0.007} & \textbf{0.786} & \textbf{0.761} & \textbf{0.087} \\

\hline

\multirow{3}{*}{\begin{tabular}[c]{@{}c@{}}Disp. Impact\\Penalty \end{tabular}}  
& CO & 0.992 & 0.992 & 0.000 & 0.671 & 0.682 & 0.011 & 0.822 & 0.716 & 0.106 \\ 
& PE & 0.920 & 0.923 & 0.002 & 0.795 & 0.808 & 0.014 & 0.851 & 0.910 & 0.058 \\
& \textbf{Avg} & \textbf{0.948} & \textbf{0.946} & \textbf{0.004} & \textbf{0.731} & \textbf{0.731} & \textbf{0.011} & \textbf{0.817} & \textbf{0.765} & \textbf{0.076} \\ 

\hline

\multirow{3}{*}{\begin{tabular}[c]{@{}c@{}}Eq. Loss \\Val\end{tabular}}  
& CO & 0.979 & 0.984 & 0.005 & 0.663 & 0.682 & 0.019 & 0.865 & 0.680 & 0.185 \\ 
& PE & 0.926 & 0.932 & 0.006 & 0.839 & 0.856 & 0.017 & 0.897 & 0.933 & 0.036 \\ 
& \textbf{Avg} & \textbf{0.937} & \textbf{0.940} & \textbf{0.004} & \textbf{0.740} & \textbf{0.745} & \textbf{0.010} & \textbf{0.830} & \textbf{0.784} & \textbf{0.062} \\ 

\hline

\multirow{3}{*}{\begin{tabular}[c]{@{}c@{}}Eq. Odds\\Val\end{tabular}}  
& CO & 0.995 & 0.995 & 0.000 & 0.670 & 0.678 & 0.008 & 0.797 & 0.705 & 0.093 \\ 
& PE & 0.945 & 0.953 & 0.008 & 0.830 & 0.845 & 0.015 & 0.896 & 0.917 & 0.021 \\ 
& \textbf{Avg} & \textbf{0.947} & \textbf{0.949} & \textbf{0.003} & \textbf{0.723} & \textbf{0.722} & \textbf{0.009} & \textbf{0.772} & \textbf{0.720} & \textbf{0.091} \\

\hline

\multirow{3}{*}{\begin{tabular}[c]{@{}c@{}}Disp. Impact\\Val\end{tabular}}  
& CO & 0.995 & 0.996 & 0.000 & 0.641 & 0.635 & 0.006 & 0.712 & 0.640 & 0.072 \\ 
& PE & 0.945 & 0.951 & 0.006 & 0.830 & 0.843 & 0.013 & 0.897 & 0.921 & 0.024 \\
& \textbf{Avg} & \textbf{0.957} & \textbf{0.959} & \textbf{0.003} & \textbf{0.733} & \textbf{0.730} & \textbf{0.008} & \textbf{0.788} & \textbf{0.765} & \textbf{0.054} \\ 

\hline

\multirow{3}{*}{\begin{tabular}[c]{@{}c@{}}Min-Max\\Loss\end{tabular}}  
& CO & 0.988 & 0.976 & 0.012 & 0.694 & 0.713 & 0.019 & 0.886 & 0.821 & 0.065 \\ 
& PE & 0.946 & 0.896 & 0.050 & 0.842 & 0.852 & 0.010 & 0.891 & 0.927 & 0.036 \\
& \textbf{Avg} & \textbf{0.958} & \textbf{0.906} & \textbf{0.052} & \textbf{0.758} & \textbf{0.762} & \textbf{0.008} & \textbf{0.840} & \textbf{0.840} & \textbf{0.049} \\ 

\bottomrule

\end{tabular}
\end{sc}
\end{small}
\end{center}
\vskip -0.25in
\end{table*}

\section{The Overfitting Problem for Fairness in Deep Learning}\label{sec:overfitting_dilemma}

\subsection{Training with Fairness Regularization}\label{subsec:fair_reg}
One common approach for mitigating unfairness 
is through imposing various fairness constraints or regularizers on the training objective. In this section, we describe the effectiveness of various regularization-based methods at improving the fairness of models trained for medical image classification and facial recognition. 

\xhdr{CheXpert} We implement three types of regularizers: equal loss, disparate impact, and equality of opportunity penalties, with the aim to achieve parity in performance (\ie~AUC scores) for both males and females.

In all previous works that apply such constraints, the experiments are either performed on linear models (\eg~SVM in~\cite{donini2018empirical}) or on small neural networks (\eg~2-layer network in~\cite{Padala2020achieving}). Under such settings, it is reasonable to assume that one can reliably measure fairness notions on the train set and expect such fairness to generalize to an unseen test set. However, as we see in our experiments, this is seldom the case with DNNs, which are highly overparametrized and can easily fit the train data~\cite{zhang2017understanding}. 
Hence, in theory, these regularizers will be ineffective since the regularizer's value will be extremely low on the train set purely as a result of overfitting.

We observe a similar trend in our empirical results reported in Table~\ref{tab:chexpert_reduced_results}.\footnote{Our full slate of results can be found in Appendix~\ref{sec:appendix_results}.} The model performs very well on the train set and thus appears to be fair, where fairness is measured by the difference in AUC value of males and females. However, when evaluated on the test set, \textit{models trained with the regularizers can be even less fair than a baseline model}. 

There is one noticeable exception in Table~\ref{tab:chexpert_reduced_results}; the equal loss regularizer is able to achieve better parity between AUC of males and females on the test set. However, we observe that this parity comes at the cost of increased disparity amongst age groups. This phenomenon is called \textit{fairness gerrymandering}, which we discuss in detail in Section~\ref{sec:gerrymandering}.

Thus, we conclude that achieving fairness via imposing constraints on the training set is challenging for DNNs. Their overparametrized nature leads to overfitting on training data and thus preventing any generalization of fairness on the test set. Moreover, overparametrization leads to a fluid decision boundary, which is prone to fairness gerrymandering.

\xhdr{Face Recognition} Since training facial recognition systems is a multi-class classification task, only an equal loss penalty can be adapted to enforce fairness in performance across genders. We find that applying a penalty on the difference in losses, even with a small coefficient, leads to improved fairness on the train set, although with a large accuracy trade-off. At the same time, the validation and test accuracy do not decrease significantly, and the accuracy gap remains close to the gap of the baseline model. Increasing the penalty size leads to a higher accuracy trade-off, yet this still does not improve fairness on the validation and test sets significantly. One possible explanation for such behavior is that the model overfits on the fairness objective to its training data. Additionally, since the validation and testing behavior of FR systems involves discarding the classification head and only using the feature extractor, one might guess that fairness on training data was embedded in the classification head but not the feature extractor.  Thus, equality of losses might be a good proxy for equality of accuracies computed using the classification head but not for accuracies computed using k-nearest neighbors in feature space on validation and test data. To test this hypothesis, we additionally compute classification accuracy on validation set and find that fairness is not improved even in this case, so we conclude that the problem is of the overfitting nature. The detailed results summarized in tables can be found in Appendix \ref{sec:appendix_results}.

\subsection{Imposing Fairness Constraints on a Holdout Set}
If the only reason that fairness constraints on the training set are ineffective were that training accuracy is so high that models appear fair regardless of their test-time behavior, then one might be able to bypass this problem by imposing the fairness penalty on a holdout set instead. 

However, as we see in our results in Table~\ref{tab:chexpert_reduced_results} and Appendix~\ref{sec:appendix_results}, this approach also fails in our settings. For both medical image classification and facial recognition tasks, in all of the cases where fairness is imposed on a holdout set, the downstream fairness on the test set deteriorates. We posit that this is because the model overfits the penalty on the validation set, which ultimately harms fairness on the test set. 

Another observation from Table~\ref{tab:chexpert_reduced_results} is that optimizing for the fairness penalties on the holdout validation set results in (slightly) higher AUC disparities on the validation set itself. This indicates that penalties such as equal loss, disparate impact and equalized odds are \textbf{bad proxies} for the fairness metric we measure here, \ie~ difference in AUCs.

\subsection{Max-Min Training}

\xhdr{Face Recognition}
For FR models, Min-Max training results in similar behavior as applying the equality of losses penalty. In particular, the training losses across genders indeed converge to similar values, however equality of losses is not achieved on the validation or test sets. As a result, misclassification rates on train data are similar for females and males, while on validation and test sets, the accuracy gap improves marginally. Therefore, we again encounter an overfitting problem with fairness being achieved on the train set but not on unseen data. 

\xhdr{CheXpert} Table~\ref{tab:chexpert_reduced_results} displays the results for Max-Min training. We observe that such a training procedure does not improve fairness on the test set.

\subsection{Adjusted Angular Margins for Face Recognition}
\label{sec:margins}

\par As we mention in \ref{sec:setup_fr}, facial recognition systems are trained using heads which increase the angular separation between classes. One way to improve fairness of the model with respect to gender is by using different angular margins during training and therefore promoting better feature discrimination for the minority class.~\citet{wang2020mitigating} applied this idea to learn a strategy for finding the optimal margins during training for coping with racial bias in face verification. 

\par To test this approach we train models with increased angular margin for females. To evaluate effectiveness of this method we follow~\citet{wang2020mitigating} and measure mean intra- and inter-class angles in addition to accuracies. The intra-class angle refers to the mean angle between the average feature vector of an identity and feature vectors of images of the same person. Inter-class angle refers to the minimal angle between the average feature vector of an identity and average feature vectors of other identities. Intuitively, we would expect that increasing angular margin for females would decrease the intra-class angle and increase the inter-class angle for female identities. 

\par Our results show that a model trained with increased angular margin for females achieves better training and validation accuracy, and intra- and inter-class separation for females. In fact, the accuracy gap on validation data drops by almost 5\% for ResNet-152 model. However, these results do not transfer to test data which consists of photos of new identities. Ultimately, the accuracy and angle metrics worsen for female groups leading to increased misclassification rates across genders. These results indicate that adjusting angular margins for mitigating unfairness leads to an overfitting problem since fairness improves only on identities that appear in the training set.  Results for this experiment can be found in Table~\ref{tab:adjusted_margins_table} and in Appendix~\ref{sec:appendix_results}.

\begin{table*}[t]
 \caption{\textbf{[Facial Recognition]}  Accuracy and intra- and inter-class angles measured for male and female images for ResNet-152 model trained with {\bf adjusted angle margins}. The numbers are measured on validation and test sets. It can be seen that 'fair' models (trained with increased angle margin for females) improve fairness on validation set, but increase the accuracy gap on test set. }
\vskip 0.15in
\begin{center}
\begin{small}
\begin{sc}
\begin{tabular}{c|c|ccc|cccc}
\toprule
\multicolumn{1}{l|}{}                                                     & Penalty  & Acc M & Acc F & Acc Gap & Intra M & Intra F & Inter M & Inter F \\
\midrule
\midrule
 \multirow{2}{*}{\begin{tabular}[c]{@{}c@{}}Validation\end{tabular}}     & baseline  & 89.0    & 82.7  & 6.3     & 33.2    & 35.6    & 68.3    & 68.3    \\
                                                                           & fair     & 87.5  & {\bf 86.1}  & 1.4     & 33.8    & {\bf 27.5}    & 68.2    & 71.2  \\
\hline
 \multirow{2}{*}{\begin{tabular}[c]{@{}c@{}}Test\end{tabular}}    & baseline  & 94.9  & 90.7  & 4.2     & 43.2    & 48.3    & 70.1    & 69.1    \\
                                                                           & fair     & 93.8  & 85.7  & 8.1     & 44.4    & 50.0      & 69.0      & 67.0      \\
 \bottomrule
 \end{tabular}
 \end{sc}

 \label{tab:adjusted_margins_table}
 \end{small}
 \end{center}
 \vskip -0.25in
 \end{table*}

\section{Fair Feature Representations Do Not Yield Fair Model Behavior}
\label{sec:representations}

\begin{table}[t]
\caption{\textbf{[Facial Recognition]} Facial recognition and gender classification accuracy for ResNet-152 model trained with {\bf sensitive information removal network on top of it}. Here $\alpha$ denotes the magnitude of adversarial penalty and for sufficiently large $\alpha$ the discriminator predicts fixed gender for all images. Gender in the first column is the gender of images penalized during training.}
\vskip 0.1in
\begin{center}
\begin{small}
\begin{sc}
\begin{tabular}{c|c|ccc|c}
\toprule
\multicolumn{1}{c|}{} & \multicolumn{1}{c|}{\multirow{2}{*}{$\alpha$}} & \multicolumn{3}{c|}{Test} & \multirow{2}{*}{\begin{tabular}[c]{@{}l@{}}Sens\\ Acc\end{tabular}} \\
& \multicolumn{1}{c|}{} & \multicolumn{1}{c}{Male} & \multicolumn{1}{c}{Female} & \multicolumn{1}{c|}{Gap} &\\ 

\midrule
\midrule

\multirow{4}{*}{\begin{tabular}[c]{@{}c@{}}Penalize\\Females\end{tabular}} & $\alpha = 0$   & 94.1  & 89.5   & 4.6   & 97.4 \\
& $\alpha = 1$  & 93.6   & 87.3   & 6.3   & 68.5  \\
& $\alpha = 2$  & 92.9   & 85.2   & 7.7   & {\bf 44.0}\\
& $\alpha = 3$  & 92.2   & 83.8   & 8.4   & 44.0 \\ 
\hline
\multirow{4}{*}{\begin{tabular}[c]{@{}c@{}}Penalize\\Males\end{tabular}}   & $\alpha = 0$  & 94.1  & 89.5  & 4.6   & 97.4  \\
& $\alpha = 1$  & 92.1  & 88.8  & 3.3 & {\bf 56.0}    \\
& $\alpha = 2$  & 89.6  & 87.2  & 2.4 & 56.0          \\
& $\alpha = 3$  & 87.7  & 85.9  & 1.8 & 56.0 \\
\bottomrule

\end{tabular}
\end{sc}

\label{tab:fair_features_table}
\end{small}
\end{center}
\vskip -0.3in
\end{table}

\par Another strategy for mitigating unfairness is through learning fair intermediate representations that are not correlated with sensitive attributes in the hope that a classifier built on top of `fair features' will yield fair predictions. A recent paper introduces SensitiveNets, a sensitive information removal network trained on top of a pre-trained feature extractor with an adversarial sensitive regularizer~\cite{morales2020sensitivenets}. Intuitively, the method learns a projection of pre-trained embeddings $\varphi(x)$ that minimizes the performance of a sensitive attribute classifier while maximizing the performance of a facial recognition system.

We apply this adversarial approach by training a sensitive information removal network for minimizing the facial recognition loss while simultaneously maximizing the probability of predicting a fixed gender class for all images. At the same time, the discriminator is trained for predicting the gender from $\varphi(x)$. Therefore, this can be formulated as a two-players game where the discriminator aims to predict the gender from features $\varphi(x)$, while the sensitive information removal network aims to output gender-independent features $\varphi(x)$ and confuse the discriminator.

\par We find that when a network is trained without adversarial regularization, the discriminator predicts gender with 97\% accuracy on the test set. Adversarial regularization with a sufficient penalty decreases the performance of a gender predictor to random, meaning that the resulting features $\varphi(x)$ are gender-independent. 

The results show that when fair features $\varphi(x)$ are obtained through manipulating male images, \eg{} when the adversarial regularizer forces the discriminator to label all images as females, the accuracy gap reduces at the expense of male accuracy. At the same time, when the adversarial regularizer manipulate female images, the model's accuracy gap only increases due to a drop in female accuracy. All results can be found in Table~\ref{tab:fair_features_table} and Appendix~\ref{sec:appendix_results}. Thus, we conclude that adversarial training decreases accuracy of the FR system on images whose feature vectors were used in the regularizer, damaging performance on all other classes.

There are principled mathematical reasons why ``fair features'' are problematic.  When the dataset is imbalanced, like Celeb-A which is majority women, the equilibrium strategy of a discriminator with no useful information is to always predict ``female.''  In this case, the feature extractor, which has the goal of fooling the discriminator, can only do so by distorting the features of men to appear female -- a strategy that disproportionately hurts accuracy of the male group.  Furthermore, in the hypothetical scenario where groups are balanced and the training process succeeds in creating features with no gender information, it becomes impossible to create a downstream classifier that assigns any label to men at a different rate than women.  This is true because if such a classifier existed, it could be used to create a better-than-random gender classifier.  In cases where the distribution of labels is different for men and women, this means that false positive and false negative rates must differ across genders.

\section{A Simple Baseline for Fairness: Label Flipping}
\label{sec:trade-off}
Many of the methods for fairness that we test above sacrifice accuracy without any gains in fairness on the testing data.  Sometimes, they sacrifice both fairness and accuracy.  Label flipping is one way to navigate the trade-off by adjusting the accuracy of individual groups~\cite{chang2020adversarial}.  This is done by randomly flipping labels in the training data of the subgroup with superior accuracy.  %
One might suppose that flipping labels will simply hurt performance, however we find that on facial recognition, this simple method can actually remedy unfairness without harming performance.

\xhdr{Face Recognition} For the facial recognition task, our models achieve higher accuracy on male images than on female images. Therefore, to decrease the accuracy gap, we randomly flip labels of a portion of male images during training. We do this by swapping male identities only with other random male identities so that female accuracy is largely preserved. We try different proportions of flipped labels: $p = 0.1, 0.3, 0.5$. Surprisingly, flipping 30\% of male labels increases female test accuracy by 0.8\% thereby decreasing the gender gap from $4.6\%$ to $3\%$.  Also, flipping half of the male labels only drops male test accuracy from $93.2\% $ to $90\%$ and results in a 1.3\% accuracy gap on test data.

\xhdr{CheXpert} On CheXpert, we randomly flip the true label of $p = 0.01, 0.05,$ and $0.1$ of samples from a particular sensitive group in each iteration. In general, we observe very unstable trends in AUC disparities when using label flipping. For example, when flipping female samples, we observe that flipping $1\%$ of samples during training results in a major drop in AUC for males, averaged over all tasks ($0.816$ to $0.700$) and a relatively lesser drop in AUC for females ($0.781$ to $0.725$). However, with $5\%$ flipped samples, we see that male AUC drops less ($0.816$ to $0.745$) than female AUC ($0.781$ to $0.697$). A similarly bizarre trend is seen when flipping male samples. AUC values of models trained on randomly flipped data are consistently lesser than the baseline model. We thus conclude that random flipping might not be the best solution to achieving fairness in this setting, since it yields unreliable trends in fairness and reliably performs worse than the baseline model. Results can be found in Appendix~\ref{sec:appendix_results}.

\section{Fairness Gerrymandering}
\label{sec:gerrymandering}

\par
\begin{figure}[t]
\vskip 0in
\centering
\includegraphics[width=\linewidth]{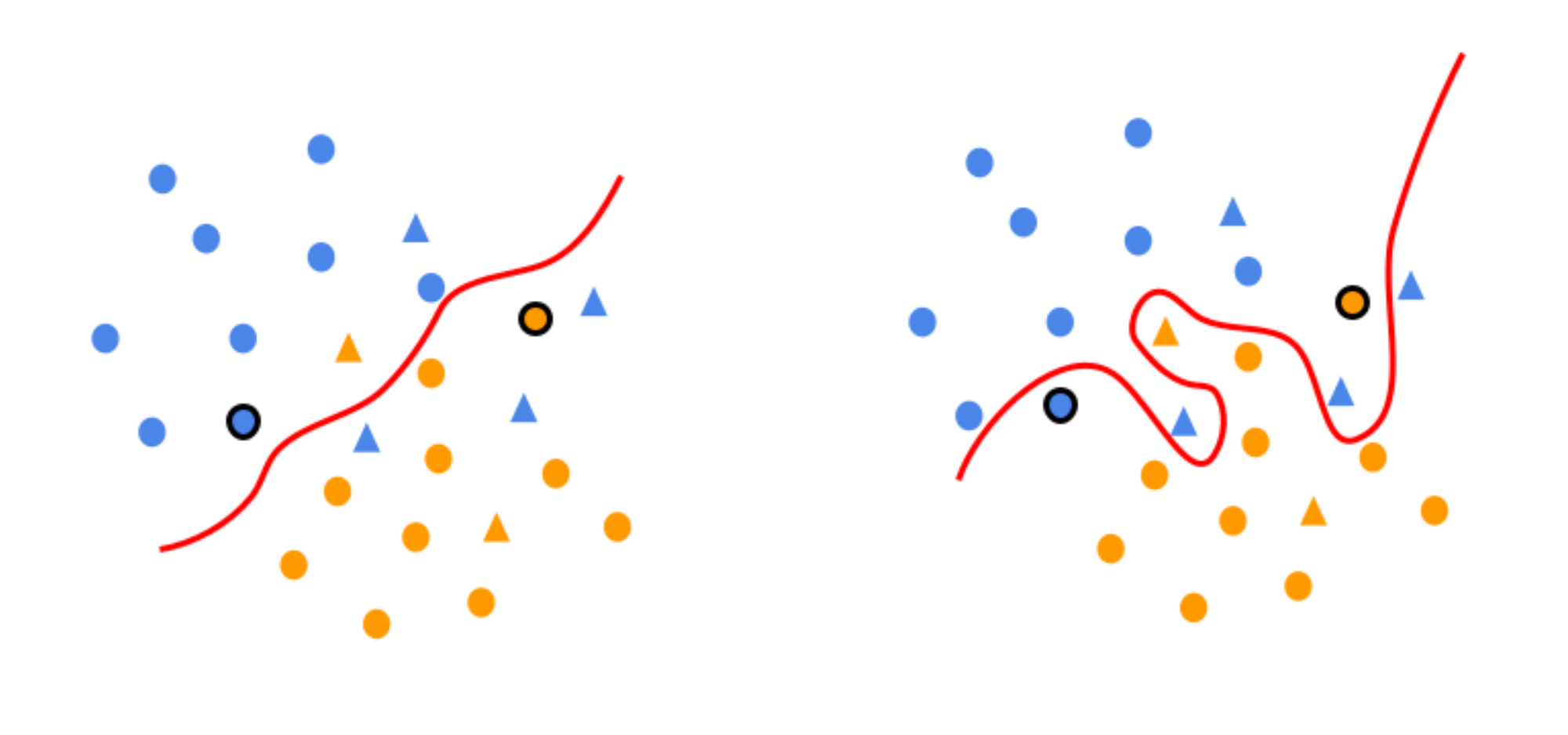}
\vspace{-10mm}
\caption{\textbf{[Schematic example of gerrymandering]} We illustrate a binary classification task with colors denoting labels, shape (circle/triangle) denoting the first sensitive feature and outline (black/transparent) denoting the second sensitive feature. The ``baseline'' classifier on the left is unfair with respect to shape. The right classifier is fair across shapes, but at the expense of points with black outline.}
\label{fig:gerrymandering_scheme}
\vskip -0.2in
\end{figure}

 Another unintended consequence of fairness interventions is fairness gerrymandering in which a model becomes more fair for one group but less fair to others ~\cite{kearns2018preventing}. Similar/related images tend to clump together in feature space.  For this reason, group-based fairness constraints that change the decision boundary are likely to induce label flips for entire groups of images with common features such as skin tone or age. Figure~\ref{fig:gerrymandering_scheme} illustrates this phenomenon.

\begin{figure}[t]
\vskip 0.2in
\centering
    \includegraphics[width=0.8\linewidth]{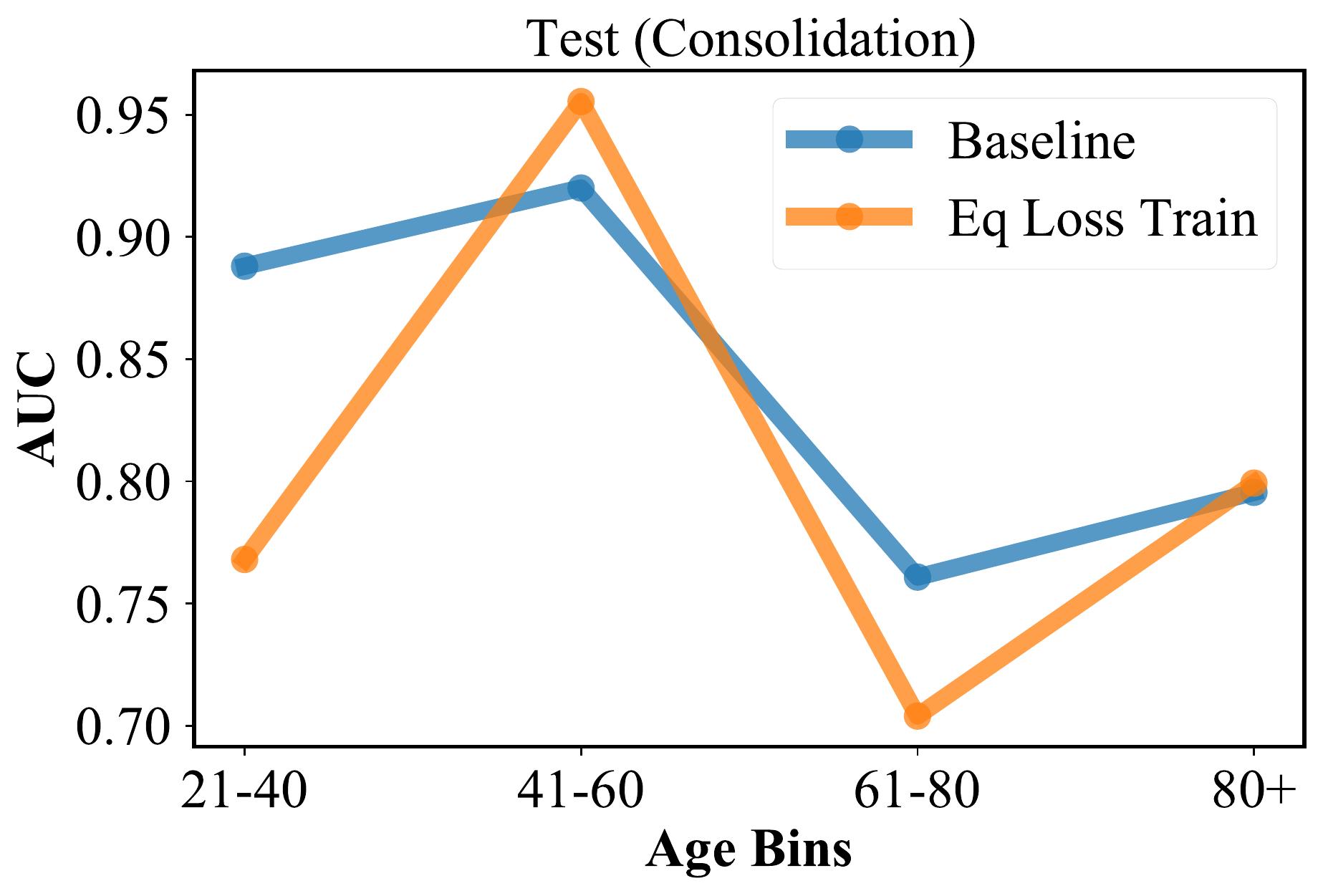}
\caption{
\textbf{[Fairness gerrymandering on CheXpert]} When the equal loss regularizer (in orange) achieves better parity along the sex demographic than the baseine training objective (in blue), we observe that it does so by making predictions more disparate along another demographic (age). In this particular example, for predicting Consolidation, the disparity between AUC of males and females reduced the most ($\approx8\%$) when using the equal loss regularizer (see Table~\ref{tab:chexpert_reduced_results}). However, as shown here, it results in higher disparities of AUC across different age groups.
}
\label{fig:chexpert_gerrymandering}
\vskip -0.2in
\end{figure}

\xhdr{CheXpert} In Table~\ref{tab:chexpert_reduced_results}, we observed better parity for models trained with an equal loss penalty than baseline training. In this section, we take a closer look at how this affected disparities across another sensitive feature, age. Consolidation (CO) is the task for which disparity in AUC across males and females reduced the most. Figure~\ref{fig:chexpert_gerrymandering} shows that this reduction in disparity across males and females induced greater disparities across age groups, which is indicated by larger differences between subsequent age groups for the ``fair'' model. We see that fairness regularization with respect to gender leads to increased unfairness with respect to age.

\xhdr{Face Recognition} We investigate if a face recognition model trained with adjusted angular margins suffers from gerrymandering.
We focus on the validation set, since this is the only data split where we observed improved fairness. 
We find that among images that flip labels relative to the baseline model, more dark-skin people flip label from correct to incorrect than dark-skin people that become classified correctly with the ``fair'' model. To test this hypothesis, we hand-label Celeb-A images in the validation subset of Celeb-A. We find that 6.3\% of images that flip to an incorrect label are dark-skin people, while only 3.8\% of images that flip to correct labels are dark-skinned. We perform a two-sample one-tail t-test and find that the difference in proportions is statistically significant with p-value $0.0078$. 

\par To further investigate this phenomenon, we test our models on race-labeled BUPT-Balancedface dataset~\cite{wang2020mitigating}. We find that the ``gender-fair'' model trained with adjusted angular margins for females is less accurate than the baseline with a significantly higher accuracy gap for Asians than for Indians ($4.6\%$ versus $2.6\%$, $p<10^{-5},$ see Table \ref{tab:gerrymandering_table}). We then compare these results to a model trained with random label flipping. Surprisingly, the model trained on corrupted data improves accuracy for all four races, but the biggest improvement happens for images of Asian people, which was the group most discriminated against by the baseline model.

\begin{table}[t]
\caption{\textbf{[Fairness gerrymandering on BUPT dataset]} The first row shows accuracies obtained with baseline model. The second row reflects performance of model trained with adjusted angle margin for females. The third row shows results for model trained with randomly flipped labels for males. For the latter two we report differences with baseline.}
\vskip 0.0in
\begin{center}
\begin{small}
\begin{sc}
\begin{tabular}{lccccc}
\toprule
Model & African & Asian & Caucasian & Indian  \\
\midrule
\midrule
Baseline & 93.3 & 90.2 & 94.1 & 94.7 \\
\hline
Margins & 88.8 & 85.6 & 90.1 & 92.1 \\
Diff & 4.5 & {\bf 4.6} & 4.0 & {\bf 2.6}\\
\hline
Flip & 93.8 & 91.7 & 95.0 & 94.4 \\
Diff & -0.5 & \bf -1.5 & -0.3 &  -0.3 \\

\bottomrule
\end{tabular}
\end{sc}
\label{tab:gerrymandering_table}
\end{small}
\end{center}
\vskip -0.25in
\end{table}

\section{Discussion}
\label{sec:discussion}

We empirically demonstrate the challenges of applying current train time algorithmic fairness interventions to DNNs. Due to overparameterization, DNNs can easily overfit the training data and thus give a false sense of fairness during training which does not generalize to the test set. In fact, we observe that adding fairness constraints via existing methods can even exacerbate unfairness on the test set. In cases where train-time fairness interventions are effective on the test set, we observe fairness gerrymandering. We posit that overparameterization makes the decision boundary of the learned neural network extremely fluid, and changing it to conform to fairness along a certain attribute can hurt fairness along another sensitive attribute. Additionally, we observe that using a holdout set to optimize fairness measures also does not yield fair outcomes on the test set, due to both overfitting and bad approximation by fairness surrogates. Our results outline some of the limitations of current train time interventions for fairness in deep learning. Evaluating other kinds of existing fairness interventions, such as pre-processing and post-processing for overparameterized models, as well as building better train time interventions are interesting avenues for future work.

\section*{Acknowledgements}

Goldstein, Cherepanova, and Goldblum were supported by the ONR MURI program, the AFOSR MURI Program, the National Science Foundation (DMS-1912866), the JP Morgan Faculty Award, and the Sloan Foundation.
Dickerson and Nanda were supported in part by NSF CAREER Award IIS-1846237, NSF D-ISN Award \#2039862, NSF Award CCF-1852352, NIH R01 Award NLM-013039-01, NIST MSE Award \#20126334, DARPA GARD \#HR00112020007, DoD WHS Award \#HQ003420F0035, and a Google Faculty Research Award. Nanda was also supported by an ERC Advanced Grant ``Foundations for Fair Social Computing'' (no. 789373).

\bibliography{example_paper}
\bibliographystyle{icml2021}

\newpage
\appendix
\section{Experimental Details}\label{sec:appendix_setup}

All code for this paper was written in Python and PyTorch.

\subsection{Medical Image Classification - CheXpert}

Table~\ref{tab:chexpert_demograhpics} shows the demographic composition of CheXpert for all the splits.

\xhdr{Data Pre-Processing} 

CheXpert contains chest x-ray images of different patients from different angles and thus does not have a fixed resolution. We resize each image to 256x256. We use the same pre-processing pipeline found in \url{https://github.com/jfhealthcare/Chexpert}. We add a Gaussian blur with $\sigma=3$ and mean normalize each image with a mean of $128.0$ and standard deviation of $64.0$.

\xhdr{Data Augmentation} 

We augment data by duplicating each image with a random affine transformation with rotations between $-15$ and $15$ degrees, a vertical and horizontal translation of $0.05$, scaling between $0.95$ and $1.05$. We fill the areas outside the transformed region with gray color (128 RGB value). Pre-processing steps are applied to each augmented image.

\xhdr{Optimizer and Hyperparameters} 

We use a batch size of 56 for all our experiments. For some of the label flip experiments, we needed to increase the batch size to 200 so as to find a non-zero number of samples to flip. Each model was trained for 20 epochs. We used the Adam optimizer with a learning rate of $0.0001$ and learning rate drops by a factor of $0.1$ every epoch. This is the same as done by the authors of \url{https://github.com/jfhealthcare/Chexpert}. For the experiments where we added a fairness regularizer (Eq~\ref{eq:eq_loss},~\ref{eq:eq_odds},~\ref{eq:di}), we need to choose an additional hyperparameter $\alpha$. We try a range of reasonable values and report results for the best $\alpha$ in the main paper. We also show results for other $\alpha$ values in Appendix~\ref{sec:appendix_results}, however we observe that higher values can interfere with the usual training objective.

\xhdr{Hardware} 

All experiments for CheXpert were run on a machine with 756GB of RAM, 2 NVIDIA Tesla V100 GPUs (32GB memory per GPU) and 32 CPU cores. It took about 12-14 hours to fully train one CheXpert model.

\begin{table}[h]
\caption{Distribution of males and females across CheXpert splits.}
\label{tab:chexpert_demograhpics}
\vskip 0.15in
\begin{center}
\begin{small}
\begin{sc}
\begin{tabular}{lccc}
\toprule
 & Train & Validation & Test \\
\midrule
\midrule
Male & $61\%$ & $53\%$ & $55\%$ \\
Female & $39\%$ & $47\%$ & $45\%$ \\ 
\bottomrule
\end{tabular}
\end{sc}
\end{small}
\end{center}
\vskip -0.1in
\end{table}

\subsection{Face Recognition}
{\bf Training routine}

\par We train facial recognition systems with ResNet-18 and ResNet-152 backbones and CosFace head using Focal Loss for 120 epochs with a batch size of 512. We utilize the SGD optimizer with a momentum of 0.9, weight decay of 5e-4 and learning rate of 0.1, and we drop the learning rate by a factor of 10 at epochs 35, 65 and 95. All images used for training contain aligned faces re-scaled to $112\times 112$. During training we use random horizontal flip data augmentation. 

For training routines, we modify the code from publicly available github repository face.evoLVe.PyTorch \footnote{https://github.com/ZhaoJ9014/face.evoLVe.PyTorch}.

{\bf Training with fairness constraints}

\par For facial recognition, we only apply an equal loss penalty, that is the absolute value of the difference between focal losses computed on male and female images in a batch. When regularization is imposed on training data, the same images are used to compute the classification loss and fairness penalty. When regularization is imposed on a holdout set, 10\% of training images are kept for enforcing the fairness penalty and are not used in the recognition objective. 

{\bf Adjusted Angular Margins}

\par Below, we provide the loss for CosFace:
$$L_C = \frac{1}{N} \sum_{i=1}^N -\log \frac{e^{s(\cos(\theta_{y_i,i})-m)}}{e^{s(\cos(\theta_{y_i,i})-m)} + \sum_{i=1, i\ne y}^{n} e^{s\cos (\theta_{j,i})}},$$
where $x_i$ is the feature vector from the i-th sample from class $y^{(i)}$. $W$ denotes the weight matrix of the last layer. Then, $W_j$ is the j-th column of $W$ and $\theta_{ij}$ denotes the angle between $W_i$ and $x_j$. A fixed parameter $m \ge 0$ controls the magnitude of the cosine margin.

\par When trained regularly, $m=0.35$ is used for both female and male images. When angular margin is adjusted for females, $m = 0.75$ is used for females and $m=0.35$ for males. 

{\bf Sensitive Information Removal Network }

\par We denote the parameters of the sensitive information removal network (SIRN) as $w$ and parameters of the sensitive classifier on top of it as $w_s$. SIRN takes as an input pre-trained embedding $x_i$ of an image $i$ from identity $y_i$ and sensitive group $s_i$ (gender) and outputs its projection $\varphi(x)$. The modified embedding is then fed into the CosFace head which outputs the logits for identities and into the sensitive head that outputs logits for genders. SIRN consist of 4 linear layers with ReLU-nonlinearities and sensitive head consist of 3 linear layers with ReLU-nonlinearities. Let $L_{FR}$ denote the focal loss for the facial recognition task and $L_{S}$ denote the cross-entropy loss for the sensitive attribute classification task.  

Then, the optimization objective for the problem is
$$
\min_{w} \frac{1}{N}\sum_i L_{FR} (\varphi(x_i), y_i) + \alpha\log(1+ |0.9 - P_s(s|\varphi(x_i))|)
$$
$$
\min_{w_s} \frac{1}{N}\sum_i L_{S} (\varphi(x_i), s_i),
$$

where $s$ is a fixed sensitive group (fixed gender), and $\alpha$ is the magnitude of the adversarial regularization term. $P_s(s|\varphi(x_i)$ denotes the sensitive head logit corresponding to gender $s$. Therefore, the first objective minimizes the facial recognition loss while simultaneously maximizing the probability of predicting a fixed gender class for all images. At the same time, the second objective minimizes the classification loss of the sensitive attribute classifier.

\section{Additional Results}\label{sec:appendix_results}

\xhdr{Face Recognition} Tables~\ref{tab:rn18_celeba} and~\ref{tab:rn152_celeba} show results for Face Recognition tasks. We also report results for additional hyperparameter values here.

\begin{figure*}[t]
\vskip 0.2in
\centering
\begin{minipage}{.32\linewidth}
\centering
\includegraphics[width=\linewidth]{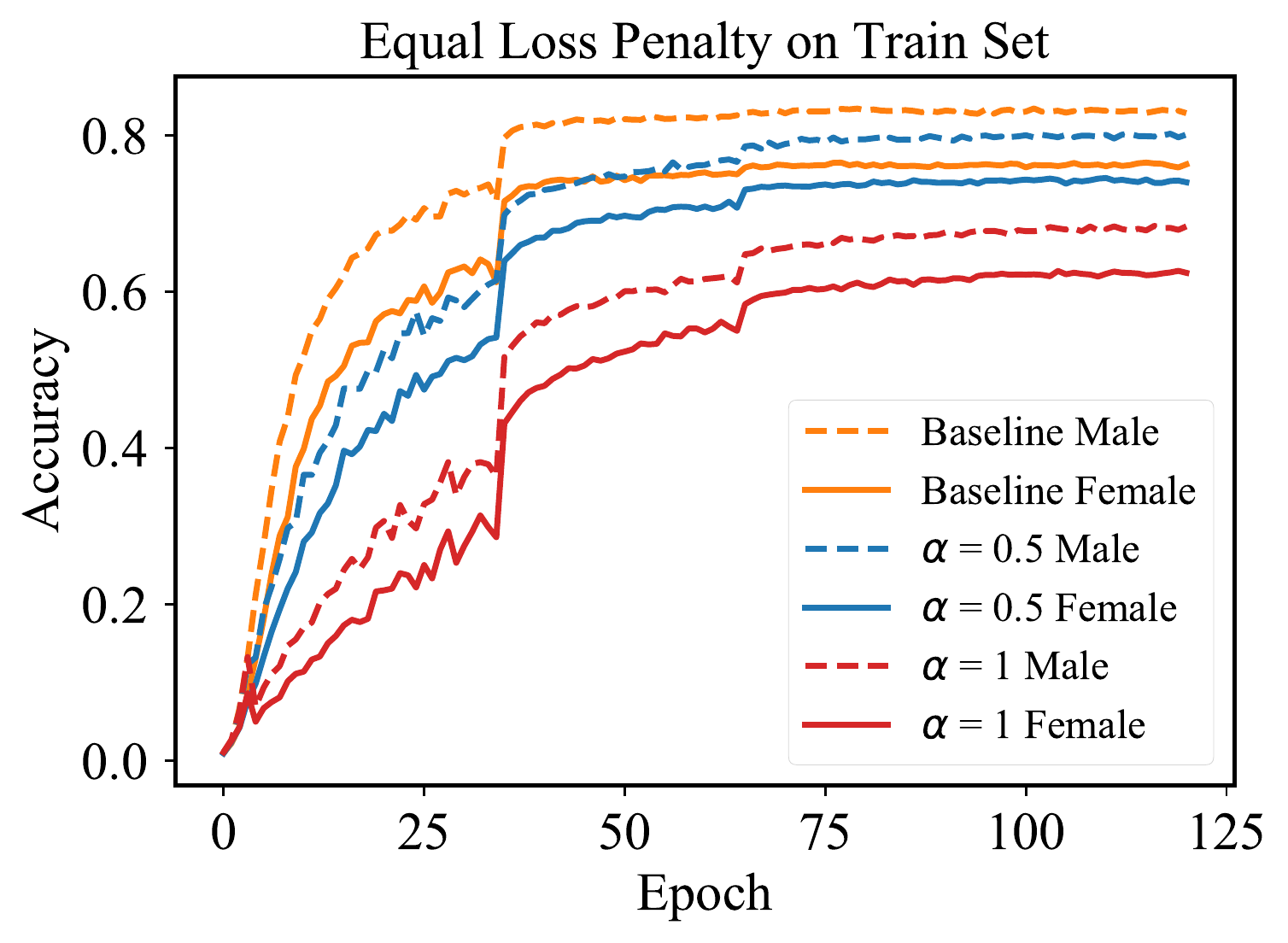}
\end{minipage}
\begin{minipage}{.32\linewidth}
\centering
\includegraphics[width=\linewidth]{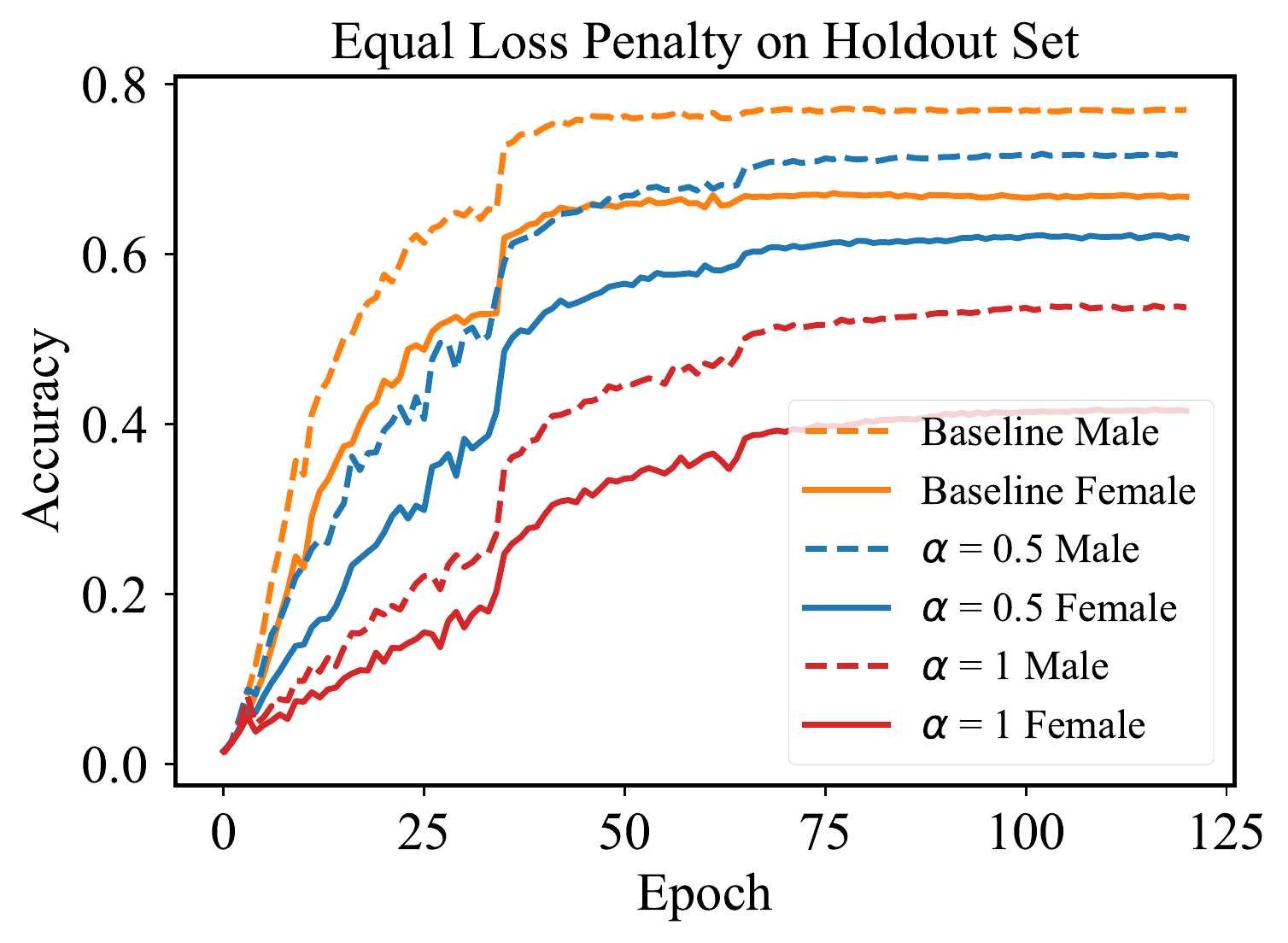}
\end{minipage}

\begin{minipage}{.32\linewidth}
\centering
\includegraphics[width=\linewidth]{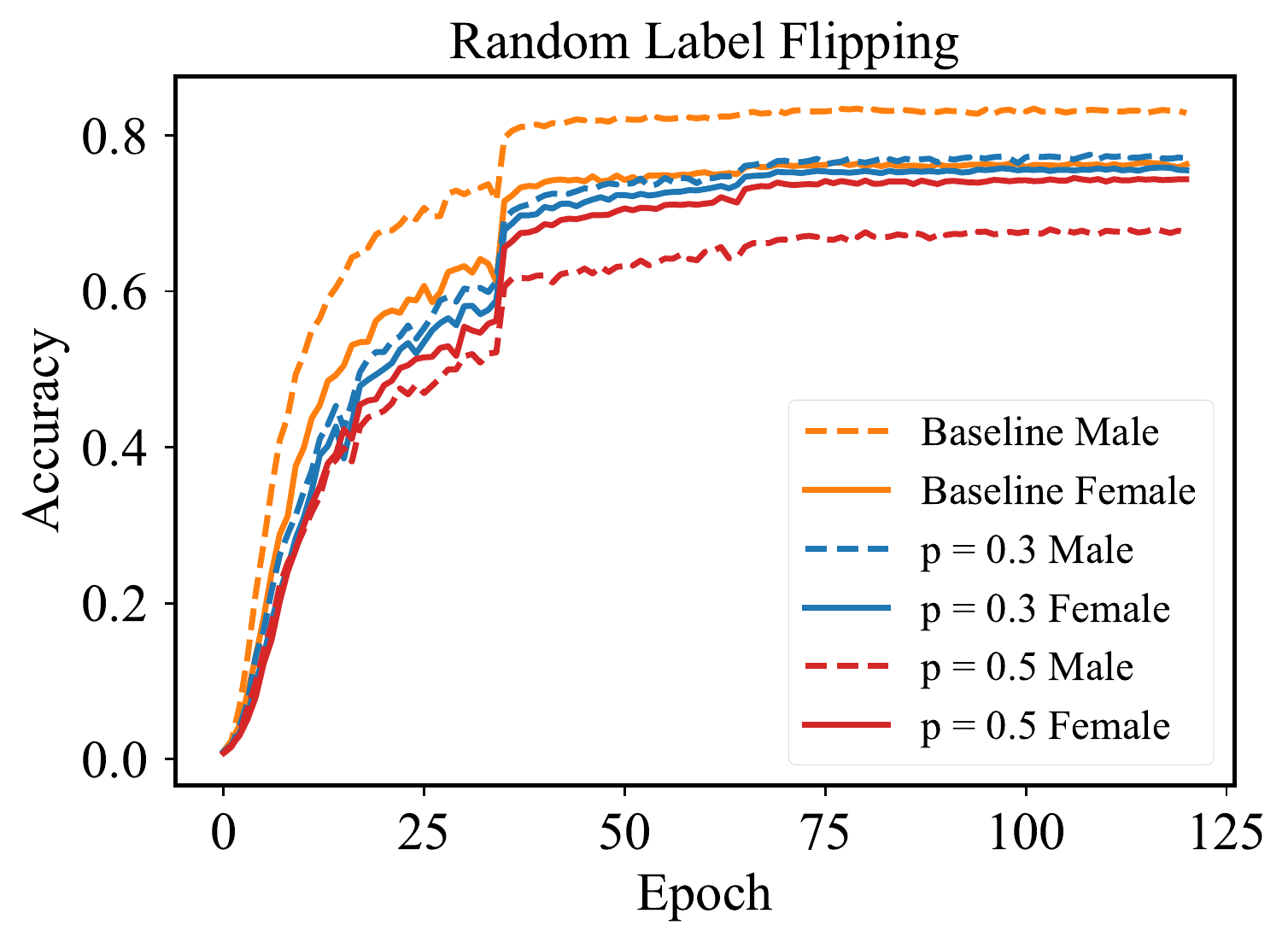}
\end{minipage}
\begin{minipage}{.32\linewidth}
\centering
\includegraphics[width=\linewidth]{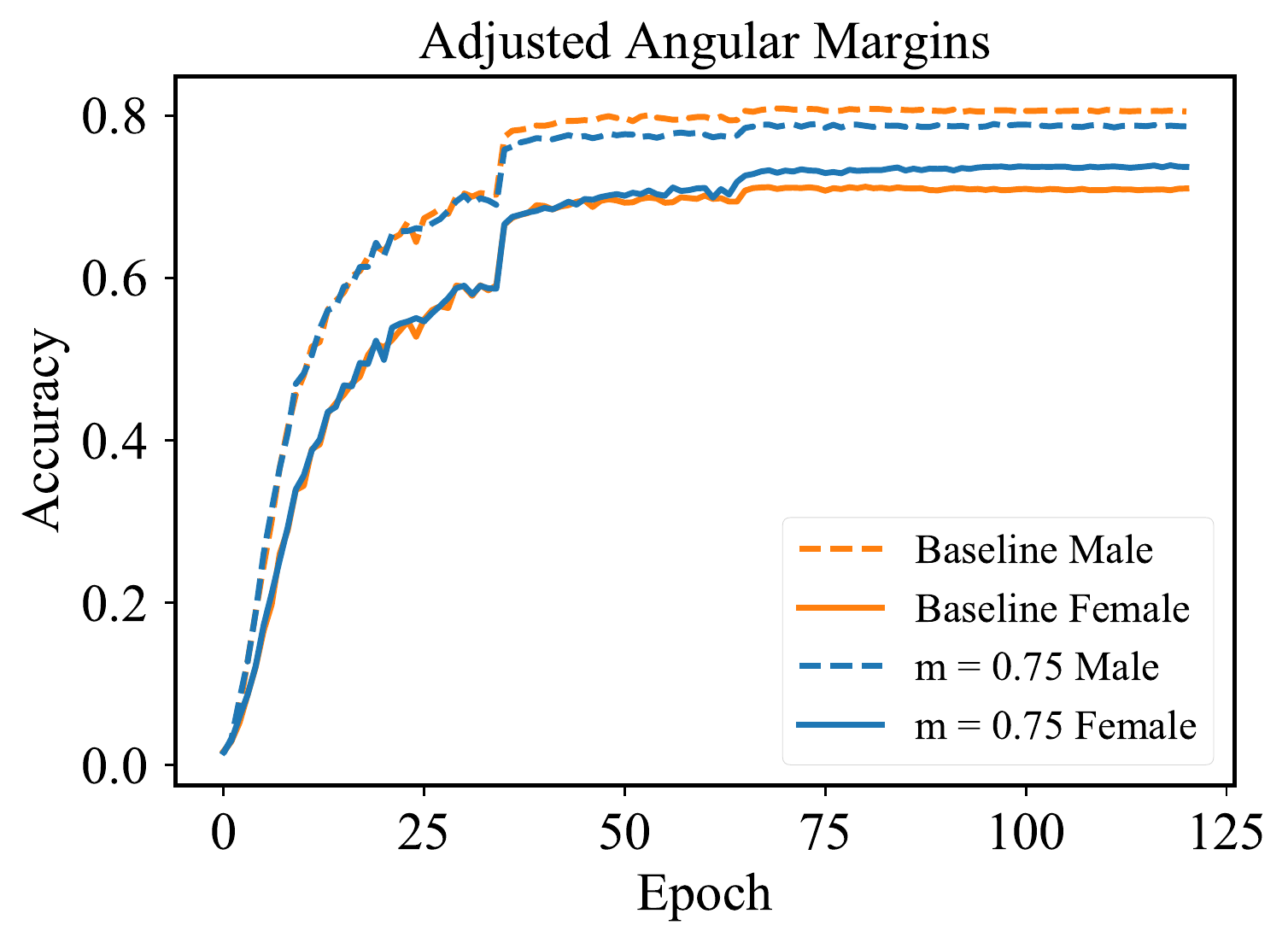}
\end{minipage}

\begin{minipage}{.32\linewidth}
\centering
\includegraphics[width=\linewidth]{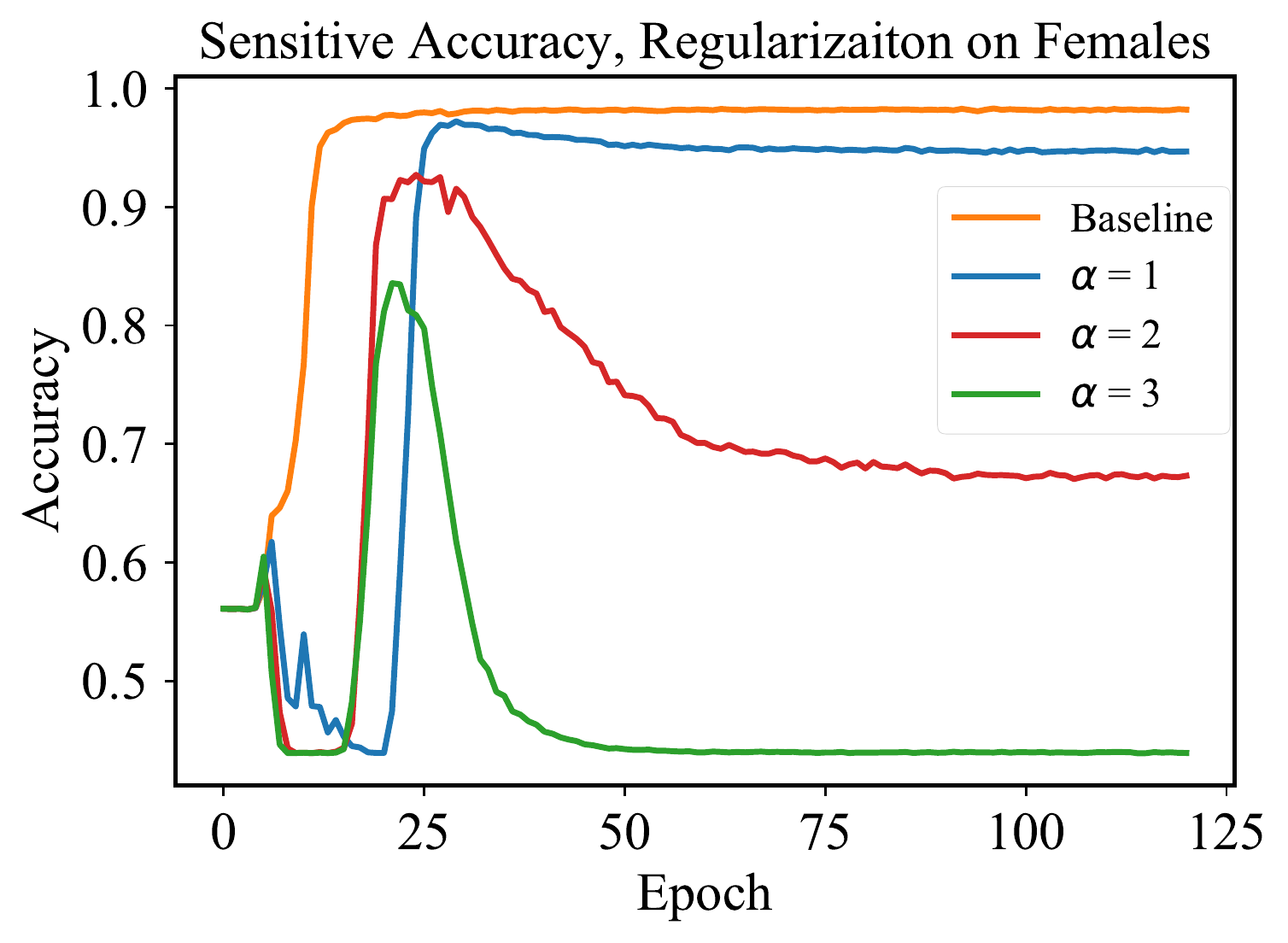}
\end{minipage}
\begin{minipage}{.32\linewidth}
\centering
\includegraphics[width=\linewidth]{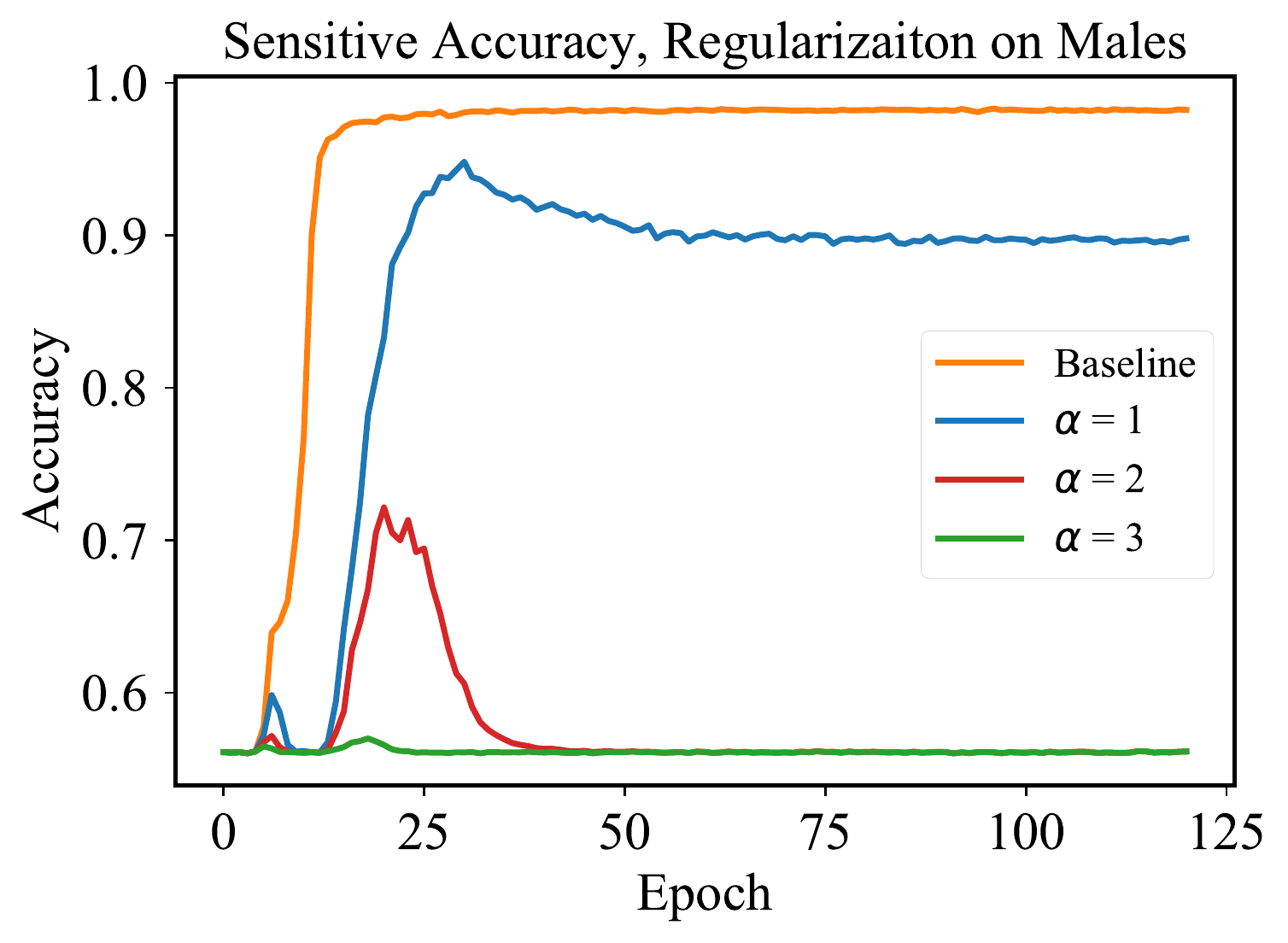}
\end{minipage}

\begin{minipage}{.45\linewidth}
\centering
\includegraphics[width=\linewidth]{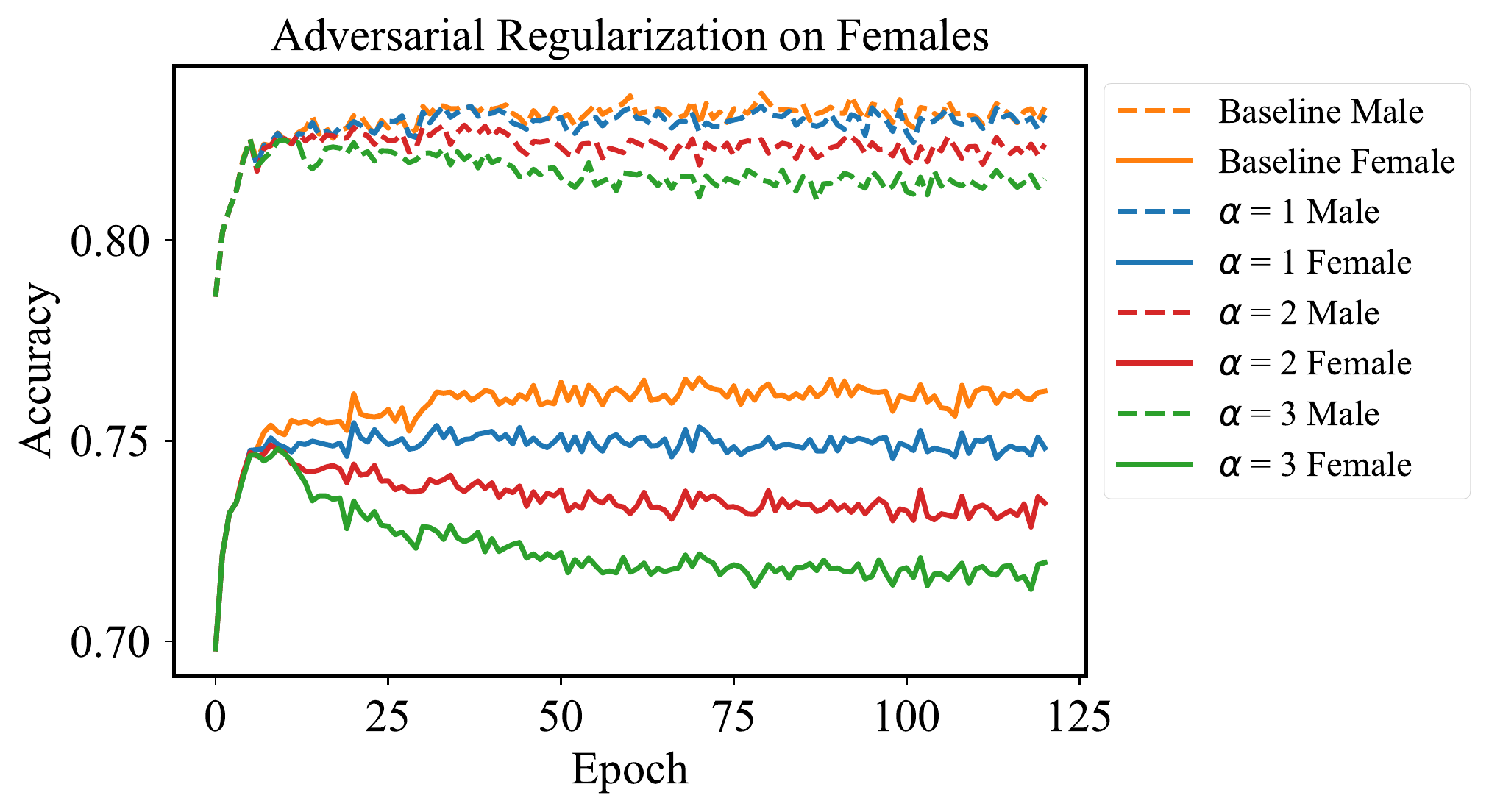}
\end{minipage}
\begin{minipage}{.45\linewidth}
\centering
\includegraphics[width=\linewidth]{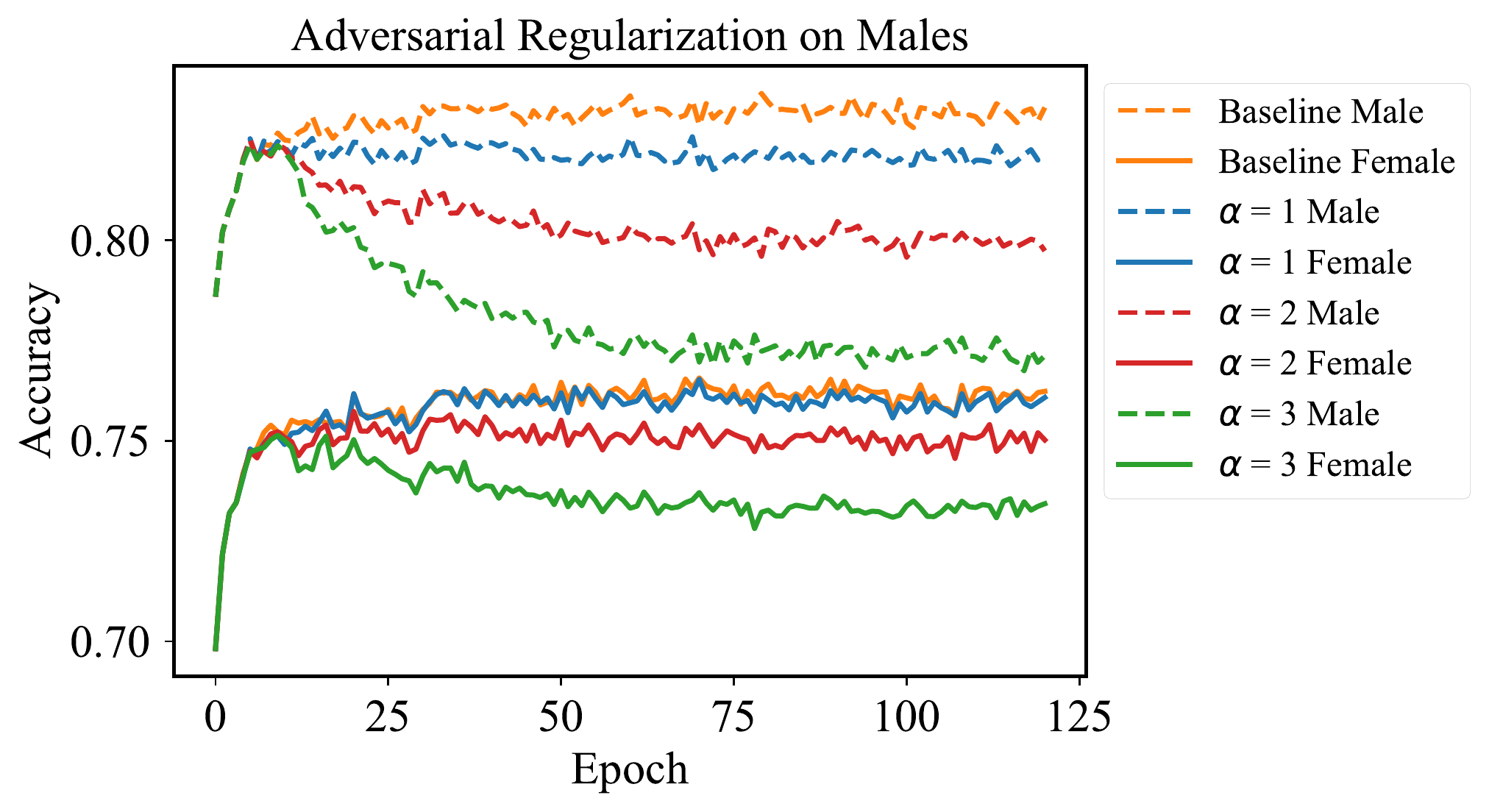}
\end{minipage}

\caption{{\bf [Training Curves for Facial Recognition]} First row: Validation accuracy of facial recognition models trained with the equal loss penalty imposed on train (1) and holdout data (2). Second row: Validation accuracy of models trained with random label flipping (3) and adjusted angular margins (4). Third row: Accuracy of the sensitive classifier predicting gender for an adversarial penalty imposed on females (5) and males (6). Fourth row: Validation accuracy of models trained with adversarial regularization imposed on females (7) and males (8). Solid lines denote accuracy of the model for females and dashed lines denotes accuracy for males. All curves are for ResNet-18 based models. }
\label{fig:celeba}
\vskip -0.2in
\end{figure*}

\xhdr{CheXpert} Tables~\ref{tab:chexpert_penalties},~\ref{tab:chexpert_val},~\ref{tab:chexpert_minmax}, and~\ref{tab:chexpert_random_flip} show results for all CheXpert tasks. We also report results for additional hyperparameter values here. Figure \ref{fig:celeba} illustrates training curves for facial recognition models.

\begin{table*}[t]
\caption{\textbf{[Face Recognition ResNet-18]} Performance of models trained with different training schemes designed for mitigating disparity in misclassification rates between males and females. All models have ResNet-18 backbone and CosFace head. The first column refers to the training scheme used, penalty indicates the size penalty coefficient. The train accuracy is computed in a classification manner, while validation and test accuracies are computed in the 1-nearest neighbors sense. The gap subcolumn refers to the difference between male and female accuracies. For the Fair Features training scheme, ``gender'' refers to the subgroup of images used in adversarial regularization during training.}
\vskip 0.15in
\begin{center}
\begin{small}
\begin{sc}
\begin{tabular}{c|c|ccc|ccc|ccc}
\toprule
\multirow{2}{*}{Scheme}                                                                 & \multirow{2}{*}{Penalty} & \multicolumn{3}{c|}{Train} & \multicolumn{3}{c|}{Validation} & \multicolumn{3}{c}{Test} \\
                                                                                        &                          & Male   & Female   & Gap    & Male     & Female     & Gap     & Male   & Female   & Gap  \\ 
\midrule
\midrule
\multirow{4}{*}{\begin{tabular}[c]{@{}c@{}}Equal Losses\\ Penalty\\ on train set\end{tabular}}   & baseline                 & 97.7   & 94.7     & 3.0    & 84.4     & 77.1       & 7.3     & 93.2   & 88.6     & 4.6  \\
                                                                                        & $\alpha = 0.5$           & 79.7   & 80.2     & -0.5   & 80.6     & 75         & 5.6     & 93.4   & 89.1     & 4.3  \\
                                                                                        & $\alpha = 1$             & 32.8   & 34.1     & -1.3   & 69.4     & 63.8       & 5.6     & 90.9   & 86.9     & 4    \\
                                                                                        & $\alpha = 3$             & 0      & 0        & 0      & 0.8      & 0.7        & 0.1     & 7.1    & 5.9      & 1.2  \\ \hline
\multirow{4}{*}{\begin{tabular}[c]{@{}c@{}}Equal Losses\\ Penalty\\ on holdout set\end{tabular}} & baseline                 & 97.1   & 93.8     & 3.3    & 81.8     & 73.5       & 8.3     & 92.7   & 87.9     & 4.8  \\
                                                                                        & $\alpha = 0.5$           & 75.1   & 62.1     & 13     & 76.7     & 69.4       & 7.3     & 92.4   & 88.1     & 4.3  \\
                                                                                        & $\alpha = 1$             & 31.3   & 21.5     & 9.8    & 60       & 50.3       & 9.7     & 87.1   & 81.1     & 6    \\
                                                                                        & $\alpha = 3$             & 0      & 0        & 0      & 2.2      & 1.6        & 0.6     & 16     & 12.8     & 3.2  \\ \hline
\multirow{4}{*}{\begin{tabular}[c]{@{}c@{}}Random \\ Labels Flipping\end{tabular}}          & baseline                 & 97.7   & 94.7     & 3.0    & 84.4     & 77.1       & 7.3     & 93.2   & 88.6     & 4.6  \\
                                                                                        & $p=0.1$                  & 94.6   & 94.2     & 0.4    & 82.5     & 77.1       & 5.4     & 93.6   & 89       & 4.6  \\
                                                                                        & $p=0.3$                  & 73.6   & 89.8     & -16.2  & 78.2     & 76.1       & 2.1     & 92.4   & 89.4     & 3    \\
                                                                                        & $p=0.5$                  & 25.3   & 88.9     & -63.6  & 68.6     & 74.9       & -6.3    & 90     & 88.7     & 1.3  \\ \hline
\multirow{2}{*}{\begin{tabular}[c]{@{}c@{}}Adjusted\\ Angle Margins\end{tabular}}       & baseline                 & 97.7   & 94.7     & 3.0    & 84.4     & 77.1       & 7.3     & 93.2   & 88.6     & 4.6  \\
                                                                                        & fair                     & 94.5   & 91.8     & 2.7    & 83.2     & 79.3       & 3.9     & 92.9   & 87.3     & 5.6  \\ \hline
\multirow{2}{*}{Min-Max}                                                                & baseline                 & 97.7   & 94.7     & 3.0    & 84.4     & 77.1       & 7.3     & 93.2   & 88.6     & 4.6  \\
                                                                                        & fair                     & 78.0   & 78.0     & 0.0    & 80.5     & 74.6       & 5.9     & 93.3   & 89.1     & 4.2  \\ \hline
\multirow{4}{*}{\begin{tabular}[c]{@{}c@{}}Fair\\  Features\\ (Females)\end{tabular}}   & baseline                 & 98.5   & 96.6     & 1.9    & 84       & 76.8       & 7.2     & 92     & 87.4     & 4.6  \\
                                                                                        & $\alpha = 1$             & 98.2   & 96.0     & 2.2    & 83.6     & 75.4       & 8.2     & 91.8   & 86.2     & 5.6  \\
                                                                                        & $\alpha = 2$             & 97.4   & 94.5     & 2.9    & 83       & 74         & 9       & 91.5   & 85.1     & 6.4  \\
                                                                                        & $\alpha = 3$             & 95.0   & 91.5     & 3.5    & 82.1     & 72.7       & 9.4     & 91.1   & 83.5     & 7.6  \\ \hline
\multirow{4}{*}{\begin{tabular}[c]{@{}c@{}}Fair\\  Features\\ (Males)\end{tabular}}     & baseline                 & 98.5   & 96.6     & 1.9    & 84       & 76.8       & 7.2     & 92     & 87.4     & 4.6  \\
                                                                                        & $\alpha = 1$             & 98.1   & 96.4     & 1.7    & 82.8     & 76.6       & 6.2     & 91.2   & 86.9     & 4.3  \\
                                                                                        & $\alpha = 2$             & 95.9   & 94.1     & 2.8    & 80.7     & 75.7       & 5       & 89.6   & 86.4     & 3.2  \\
                                                                                        & $\alpha = 3$             & 91.6   & 88.5     & 3.1    & 77.8     & 74.1       & 3.7     & 87.8   & 85.1     & 2.7  \\
\bottomrule
\end{tabular}
\end{sc}
\label{tab:rn18_celeba}
\end{small}
\end{center}
\vskip -0.1in
\end{table*}

\begin{table*}[t]
\caption{ \textbf{[Face Recognition ResNet-152]} Performance of models trained with different training schemes designed for mitigating disparity in misclassification rates between males and females. All models have ResNet-152 backbone and CosFace head. The first column refers to the training scheme used, penalty indicates the size penalty coefficient. The train accuracy is computed in a classification manner, while validation and test accuracies are computed in the 1-nearest neighbors sense. The gap subcolumn refers to the difference between male and female accuracies. For the Fair Features training scheme, gender refers to the subgroup of images used in adversarial regularization during training.  }
\vskip 0.15in
\begin{center}
\begin{small}
\begin{sc}
\begin{tabular}{c|c|ccc|ccc|ccc}
\toprule
\multirow{2}{*}{Scheme}                                                               & \multirow{2}{*}{Penalty} & \multicolumn{3}{c|}{Train} & \multicolumn{3}{c|}{Validation} & \multicolumn{3}{c}{Test} \\ 
                                                                                      &                          & Male   & Female   & Gap    & Male     & Female     & Gap     & Male   & Female   & Gap  \\ 
\midrule
\midrule
\multirow{3}{*}{\begin{tabular}[c]{@{}c@{}}Focal Penalty\\ on train set\end{tabular}} & baseline                 & 99.3   & 98.2     & 1.1    & 89       & 82.7       & 6.3     & 94.9   & 90.7     & 4.2  \\
                                                                                      & $\alpha = 0.5$           & 90.0   & 91.3     & -1.3   & 88       & 82.5       & 5.5     & 95.5   & 91.6     & 3.9  \\
                                                                                      & $\alpha = 1$             & 46     & 47.6     & -1.6   & 78.1     & 72.3       & 5.8     & 93.3   & 89.9     & 3.4  \\ \hline
\multirow{3}{*}{\begin{tabular}[c]{@{}c@{}}Random \\ Labels Flip\end{tabular}}        & baseline                 & 99.3   & 98.2     & 1.1    & 89       & 82.7       & 6.3     & 94.9   & 90.7     & 4.2  \\
                                                                                      & $p=0.3$                  & 84.8   & 96.4     & -11.6  & 85.3     & 83.4       & 1.9     & 94.9   & 91.8     & 3.1  \\
                                                                                      & $p=0.5$                  & 35.0   & 95.4     & -60.4  & 79.1     & 82.8       & -3.7    & 93.2   & 91.4     & 1.8  \\ \hline
\multirow{2}{*}{\begin{tabular}[c]{@{}c@{}}Adjusted\\ Angle Margins\end{tabular}}   & baseline                 & 99.3   & 98.2     & 1.1    & 89       & 82.7       & 6.3     & 94.9   & 90.7     & 4.2  \\
                                                                                      & fair                     & 99.3   & 98.9     & 0.4    & 87.5     & 86.1       & 1.4     & 93.8   & 85.7     & 8.1  \\ \hline
\multirow{4}{*}{\begin{tabular}[c]{@{}c@{}}Fair\\  Features\\ (Females)\end{tabular}} & baseline                 & 99.5   & 98.7     & 0.8    & 88.3     & 82         & 6.3     & 94.1   & 89.5     & 4.6  \\
                                                                                      & $\alpha = 1$             & 98.8   & 97.4     & 1.4    & 87.5     & 79.6       & 7.9     & 93.6   & 87.3     & 6.3  \\
                                                                                      & $\alpha = 2$             & 96.6   & 94.7     & 1.9    & 86.8     & 77.3       & 9.5     & 92.9   & 85.2     & 7.7  \\
                                                                                      & $\alpha = 3$             & 93.8   & 92.0     & 1.8    & 85.8     & 75.3       & 10.5    & 92.2   & 83.8     & 8.4  \\ \hline
\multirow{4}{*}{\begin{tabular}[c]{@{}c@{}}Fair\\  Features\\ (Males)\end{tabular}}   & baseline                 & 99.5   & 98.7     & 0.8    & 88.3     & 82         & 6.3     & 94.1   & 89.5     & 4.6  \\
                                                                                      & $\alpha = 1$             & 97.7   & 97.3     & 0.4    & 85.5     & 80.9       & 4.6     & 92.1   & 88.8     & 3.3  \\
                                                                                      & $\alpha = 2$             & 93.8   & 92.2     & 1.6    & 82.8     & 78.9       & 3.9     & 89.6   & 87.2     & 2.4  \\
                                                                                      & $\alpha = 3$             & 90.6   & 87.8     & 2.8    & 80.8     & 77.5       & 3.3     & 87.7   & 85.9     & 1.8  \\ 

\bottomrule
\end{tabular}
\end{sc}
\label{tab:rn152_celeba}
\end{small}
\end{center}
\vskip -0.1in
\end{table*}

\begin{table*}[t]
\caption{\textbf{[CheXpert - training with fairness penalties]} Results for all 5 CheXpert tasks: Cardiomegaly (CA), Edema (ED), Consolidation (CO), Atelectasis (AT) and Pleural Effusion (PE). The regularizer is optimized on the training data, and $\alpha$ here denotes the coefficient of the regularizer. Results in Table~\ref{tab:chexpert_reduced_results} (main paper) are for $\alpha=100$ on the equal loss penalty and $\alpha=1000$ on the equal odds and disparate impact penalties.}
\label{tab:chexpert_penalties}
\vskip 0.15in
\begin{center}
\begin{small}
\begin{sc}
\begin{tabular}{c|c|ccc|ccc|ccc}
\toprule
\multirow{2}{*}{Scheme} & \multirow{2}{*}{Task} & \multicolumn{3}{c|}{Train} & \multicolumn{3}{c|}{Validation} & \multicolumn{3}{c}{Test} \\
 &   & Male   & Female   & Gap    & Male     & Female     & Gap     & Male   & Female   & Gap  \\ 
\midrule
\midrule

\multirow{6}{*}{\begin{tabular}[c]{@{}c@{}}Baseline\end{tabular}} 
& CD & 0.988 & 0.990 & 0.002 & 0.826 & 0.818 & 0.007 & 0.766 & 0.739 & 0.027 \\
& ED & 0.953 & 0.952 & 0.000 & 0.780 & 0.777 & 0.002 & 0.885 & 0.862 & 0.024 \\ 
& CO & 0.996 & 0.996 & 0.000 & 0.681 & 0.687 & 0.006 & 0.896 & 0.808 & 0.088 \\ 
& AT & 0.879 & 0.883 & 0.004 & 0.637 & 0.631 & 0.007 & 0.637 & 0.570 & 0.068 \\ 
& PE & 0.936 & 0.942 & 0.006 & 0.841 & 0.854 & 0.013 & 0.895 & 0.925 & 0.030 \\
& \textbf{Avg} & \textbf{0.950} & \textbf{0.953} & \textbf{0.002} & \textbf{0.753} & \textbf{0.753} & \textbf{0.007} & \textbf{0.816} & \textbf{0.781} & \textbf{0.047} \\ 

\hline\hline

\multirow{6}{*}{\begin{tabular}[c]{@{}c@{}}Eq. Loss \\Train\\$\alpha=100$\end{tabular}} 
& CD & 0.974 & 0.975 & 0.001 & 0.833 & 0.826 & 0.007 & 0.724 & 0.715 & 0.009 \\ 
& ED & 0.912 & 0.906 & 0.007 & 0.788 & 0.788 & 0.000 & 0.925 & 0.879 & 0.046 \\ 
& CO & 0.983 & 0.981 & 0.003 & 0.699 & 0.698 & 0.000 & 0.835 & 0.819 & 0.016 \\ 
& AT & 0.766 & 0.767 & 0.001 & 0.674 & 0.672 & 0.003 & 0.863 & 0.855 & 0.008 \\ 
& PE & 0.900 & 0.906 & 0.006 & 0.850 & 0.864 & 0.015 & 0.886 & 0.946 & 0.059 \\
& \textbf{Avg} & \textbf{0.907} & \textbf{0.907} & \textbf{0.003} & \textbf{0.769} & \textbf{0.770} & \textbf{0.005} & \textbf{0.847} & \textbf{0.843} & \textbf{0.028} \\ 

\hline\hline

\multirow{6}{*}{\begin{tabular}[c]{@{}c@{}}Eq. Loss \\Train\\$\alpha=1000$\end{tabular}} 
& CD & 0.807 & 0.795 & 0.012 & 0.792 & 0.790 & 0.002 & 0.832 & 0.786 & 0.046 \\ 
& ED & 0.814 & 0.801 & 0.013 & 0.753 & 0.745 & 0.007 & 0.862 & 0.818 & 0.044 \\ 
& CO & 0.668 & 0.675 & 0.007 & 0.674 & 0.686 & 0.013 & 0.911 & 0.780 & 0.131 \\ 
& AT & 0.575 & 0.557 & 0.018 & 0.530 & 0.528 & 0.001 & 0.573 & 0.551 & 0.022 \\ 
& PE & 0.822 & 0.834 & 0.012 & 0.810 & 0.830 & 0.020 & 0.795 & 0.898 & 0.103 \\
& \textbf{Avg} & \textbf{0.737} & \textbf{0.732} & \textbf{0.012} & \textbf{0.712} & \textbf{0.716} & \textbf{0.009} & \textbf{0.794} & \textbf{0.767} & \textbf{0.069} \\ 

\hline\hline

\multirow{6}{*}{\begin{tabular}[c]{@{}c@{}}Disp. Impact\\Penalty\\$\alpha=100$\end{tabular}}  
& CD & 0.996 & 0.997 & 0.001 & 0.813 & 0.815 & 0.001 & 0.739 & 0.757 & 0.018 \\ 
& ED & 0.969 & 0.967 & 0.002 & 0.783 & 0.783 & 0.000 & 0.964 & 0.804 & 0.160 \\ 
& CO & 0.993 & 0.994 & 0.001 & 0.672 & 0.683 & 0.011 & 0.781 & 0.788 & 0.007 \\ 
& AT & 0.892 & 0.895 & 0.003 & 0.615 & 0.607 & 0.008 & 0.741 & 0.662 & 0.079 \\ 
& PE & 0.943 & 0.951 & 0.007 & 0.817 & 0.841 & 0.024 & 0.871 & 0.888 & 0.017 \\
& \textbf{Avg} & \textbf{0.959} & \textbf{0.961} & \textbf{0.003} & \textbf{0.740} & \textbf{0.746} & \textbf{0.009} & \textbf{0.819} & \textbf{0.780} & \textbf{0.056} \\ 

\hline\hline

\multirow{6}{*}{\begin{tabular}[c]{@{}c@{}}Disp. Impact\\Penalty\\$\alpha=1000$\end{tabular}}  
& CD & 0.997 & 0.997 & 0.000 & 0.804 & 0.802 & 0.003 & 0.752 & 0.668 & 0.084 \\ 
& ED & 0.969 & 0.965 & 0.004 & 0.768 & 0.762 & 0.005 & 0.910 & 0.861 & 0.050 \\ 
& CO & 0.992 & 0.992 & 0.000 & 0.671 & 0.682 & 0.011 & 0.822 & 0.716 & 0.106 \\ 
& AT & 0.865 & 0.852 & 0.012 & 0.618 & 0.598 & 0.020 & 0.750 & 0.669 & 0.081 \\ 
& PE & 0.920 & 0.923 & 0.002 & 0.795 & 0.808 & 0.014 & 0.851 & 0.910 & 0.058 \\
& \textbf{Avg} & \textbf{0.948} & \textbf{0.946} & \textbf{0.004} & \textbf{0.731} & \textbf{0.731} & \textbf{0.011} & \textbf{0.817} & \textbf{0.765} & \textbf{0.076} \\ 

\hline\hline

\multirow{6}{*}{\begin{tabular}[c]{@{}c@{}}Eq. Odds\\Penalty\\$\alpha=100$\end{tabular}}  
& CD & 0.994 & 0.995 & 0.001 & 0.805 & 0.802 & 0.003 & 0.701 & 0.703 & 0.003 \\ 
& ED & 0.945 & 0.940 & 0.005 & 0.722 & 0.716 & 0.006 & 0.841 & 0.778 & 0.064 \\ 
& CO & 0.993 & 0.995 & 0.001 & 0.644 & 0.653 & 0.009 & 0.744 & 0.803 & 0.058 \\ 
& AT & 0.860 & 0.868 & 0.008 & 0.615 & 0.608 & 0.007 & 0.697 & 0.602 & 0.095 \\ 
& PE & 0.924 & 0.930 & 0.005 & 0.811 & 0.820 & 0.009 & 0.775 & 0.864 & 0.089 \\
& \textbf{Avg} & \textbf{0.943} & \textbf{0.945} & \textbf{0.004} & \textbf{0.719} & \textbf{0.720} & \textbf{0.007} & \textbf{0.752} & \textbf{0.750} & \textbf{0.062} \\ 

\hline\hline

\multirow{6}{*}{\begin{tabular}[c]{@{}c@{}}Eq. Odds\\Penalty\\$\alpha=1000$\end{tabular}}  
& CD & 0.996 & 0.996 & 0.000 & 0.817 & 0.812 & 0.005 & 0.747 & 0.808 & 0.061 \\ 
& ED & 0.961 & 0.962 & 0.001 & 0.765 & 0.772 & 0.007 & 0.924 & 0.825 & 0.099 \\ 
& CO & 0.998 & 0.998 & 0.000 & 0.659 & 0.664 & 0.004 & 0.781 & 0.665 & 0.116 \\ 
& AT & 0.886 & 0.890 & 0.004 & 0.629 & 0.625 & 0.004 & 0.708 & 0.642 & 0.066 \\ 
& PE & 0.928 & 0.934 & 0.005 & 0.823 & 0.836 & 0.013 & 0.769 & 0.863 & 0.094 \\
& \textbf{Avg} & \textbf{0.954} & \textbf{0.956} & \textbf{0.002} & \textbf{0.739} & \textbf{0.742} & \textbf{0.007} & \textbf{0.786} & \textbf{0.761} & \textbf{0.087} \\

\hline\hline

\end{tabular}
\end{sc}
\end{small}
\end{center}
\end{table*}

\begin{table*}[t]
\caption{\textbf{[CheXpert - fairness penalties on the validation set]} Results for all 5 CheXpert tasks: Cardiomegaly (CA), Edema (ED), Consolidation (CO), Atelectasis (AT) and Pleural Effusion (PE).}
\label{tab:chexpert_val}
\vskip 0.15in
\begin{center}
\begin{small}
\begin{sc}
\begin{tabular}{c|c|ccc|ccc|ccc}
\toprule
\multirow{2}{*}{Scheme} & \multirow{2}{*}{Task} & \multicolumn{3}{c|}{Train} & \multicolumn{3}{c|}{Validation} & \multicolumn{3}{c}{Test} \\
 &  & Male   & Female   & Gap    & Male     & Female     & Gap     & Male   & Female   & Gap  \\ 
\midrule
\midrule

\multirow{6}{*}{\begin{tabular}[c]{@{}c@{}}Baseline\end{tabular}} 
& CD & 0.988 & 0.990 & 0.002 & 0.826 & 0.818 & 0.007 & 0.766 & 0.739 & 0.027 \\
& ED & 0.953 & 0.952 & 0.000 & 0.780 & 0.777 & 0.002 & 0.885 & 0.862 & 0.024 \\ 
& CO & 0.996 & 0.996 & 0.000 & 0.681 & 0.687 & 0.006 & 0.896 & 0.808 & 0.088 \\ 
& AT & 0.879 & 0.883 & 0.004 & 0.637 & 0.631 & 0.007 & 0.637 & 0.570 & 0.068 \\ 
& PE & 0.936 & 0.942 & 0.006 & 0.841 & 0.854 & 0.013 & 0.895 & 0.925 & 0.030 \\
& \textbf{Avg} & \textbf{0.950} & \textbf{0.953} & \textbf{0.002} & \textbf{0.753} & \textbf{0.753} & \textbf{0.007} & \textbf{0.816} & \textbf{0.781} & \textbf{0.047} \\ 

\hline\hline

\multirow{6}{*}{\begin{tabular}[c]{@{}c@{}}Eq. Loss \\Val\end{tabular}}  
& CD & 0.985 & 0.987 & 0.002 & 0.827 & 0.824 & 0.002 & 0.745 & 0.669 & 0.076 \\ 
& ED & 0.931 & 0.930 & 0.001 & 0.746 & 0.735 & 0.011 & 0.885 & 0.877 & 0.008 \\ 
& CO & 0.979 & 0.984 & 0.005 & 0.663 & 0.682 & 0.019 & 0.865 & 0.680 & 0.185 \\ 
& AT & 0.863 & 0.868 & 0.005 & 0.626 & 0.629 & 0.002 & 0.759 & 0.763 & 0.004 \\ 
& PE & 0.926 & 0.932 & 0.006 & 0.839 & 0.856 & 0.017 & 0.897 & 0.933 & 0.036 \\
& \textbf{Avg} & \textbf{0.937} & \textbf{0.940} & \textbf{0.004} & \textbf{0.740} & \textbf{0.745} & \textbf{0.010} & \textbf{0.830} & \textbf{0.784} & \textbf{0.062} \\ 

\hline\hline

\multirow{6}{*}{\begin{tabular}[c]{@{}c@{}}Eq. Odds\\Val\end{tabular}}  
& CD & 0.994 & 0.996 & 0.002 & 0.776 & 0.774 & 0.002 & 0.637 & 0.715 & 0.078 \\ 
& ED & 0.936 & 0.934 & 0.003 & 0.729 & 0.721 & 0.009 & 0.858 & 0.699 & 0.159 \\ 
& CO & 0.995 & 0.995 & 0.000 & 0.670 & 0.678 & 0.008 & 0.797 & 0.705 & 0.093 \\ 
& AT & 0.867 & 0.867 & 0.001 & 0.608 & 0.595 & 0.013 & 0.671 & 0.565 & 0.106 \\ 
& PE & 0.945 & 0.953 & 0.008 & 0.830 & 0.845 & 0.015 & 0.896 & 0.917 & 0.021 \\
& \textbf{Avg} & \textbf{0.947} & \textbf{0.949} & \textbf{0.003} & \textbf{0.723} & \textbf{0.722} & \textbf{0.009} & \textbf{0.772} & \textbf{0.720} & \textbf{0.091} \\

\hline\hline

\multirow{6}{*}{\begin{tabular}[c]{@{}c@{}}Disp. Impact\\Val\end{tabular}}  
& CD & 0.997 & 0.997 & 0.000 & 0.795 & 0.788 & 0.007 & 0.691 & 0.708 & 0.018 \\ 
& ED & 0.972 & 0.973 & 0.001 & 0.768 & 0.760 & 0.008 & 0.853 & 0.891 & 0.038 \\ 
& CO & 0.995 & 0.996 & 0.000 & 0.641 & 0.635 & 0.006 & 0.712 & 0.640 & 0.072 \\ 
& AT & 0.874 & 0.879 & 0.005 & 0.633 & 0.626 & 0.007 & 0.786 & 0.668 & 0.118 \\ 
& PE & 0.945 & 0.951 & 0.006 & 0.830 & 0.843 & 0.013 & 0.897 & 0.921 & 0.024 \\
& \textbf{Avg} & \textbf{0.957} & \textbf{0.959} & \textbf{0.003} & \textbf{0.733} & \textbf{0.730} & \textbf{0.008} & \textbf{0.788} & \textbf{0.765} & \textbf{0.054} \\ 

\hline\hline

\end{tabular}
\end{sc}
\end{small}
\end{center}
\end{table*}

\begin{table*}[t]
\caption{\textbf{[CheXpert - minmax training]} Results for all 5 CheXpert tasks: Cardiomegaly (CA), Edema (ED), Consolidation (CO), Atelectasis (AT) and Pleural Effusion (PE). We see that minmax training does not lead to increased fairness on the test set.}
\label{tab:chexpert_minmax}
\vskip 0.15in
\begin{center}
\begin{small}
\begin{sc}
\begin{tabular}{c|c|ccc|ccc|ccc}
\toprule
\multirow{2}{*}{Scheme} & \multirow{2}{*}{Task} & \multicolumn{3}{c|}{Train} & \multicolumn{3}{c|}{Validation} & \multicolumn{3}{c}{Test} \\
 &  & Male   & Female   & Gap    & Male     & Female     & Gap     & Male   & Female   & Gap  \\ 
\midrule
\midrule

\multirow{6}{*}{\begin{tabular}[c]{@{}c@{}}Baseline\end{tabular}} 
& CD & 0.988 & 0.990 & 0.002 & 0.826 & 0.818 & 0.007 & 0.766 & 0.739 & 0.027 \\
& ED & 0.953 & 0.952 & 0.000 & 0.780 & 0.777 & 0.002 & 0.885 & 0.862 & 0.024 \\ 
& CO & 0.996 & 0.996 & 0.000 & 0.681 & 0.687 & 0.006 & 0.896 & 0.808 & 0.088 \\ 
& AT & 0.879 & 0.883 & 0.004 & 0.637 & 0.631 & 0.007 & 0.637 & 0.570 & 0.068 \\ 
& PE & 0.936 & 0.942 & 0.006 & 0.841 & 0.854 & 0.013 & 0.895 & 0.925 & 0.030 \\
& \textbf{Avg} & \textbf{0.950} & \textbf{0.953} & \textbf{0.002} & \textbf{0.753} & \textbf{0.753} & \textbf{0.007} & \textbf{0.816} & \textbf{0.781} & \textbf{0.047} \\ 

\hline\hline

\multirow{6}{*}{\begin{tabular}[c]{@{}c@{}}Min-Max\\Loss\end{tabular}}  
& CD & 0.992 & 0.981 & 0.011 & 0.824 & 0.820 & 0.004 & 0.771 & 0.789 & 0.018 \\ 
& ED & 0.963 & 0.904 & 0.058 & 0.774 & 0.774 & 0.000 & 0.895 & 0.838 & 0.057 \\ 
& CO & 0.988 & 0.976 & 0.012 & 0.694 & 0.713 & 0.019 & 0.886 & 0.821 & 0.065 \\ 
& AT & 0.901 & 0.771 & 0.130 & 0.657 & 0.652 & 0.004 & 0.758 & 0.826 & 0.068 \\ 
& PE & 0.946 & 0.896 & 0.050 & 0.842 & 0.852 & 0.010 & 0.891 & 0.927 & 0.036 \\
& \textbf{Avg} & \textbf{0.958} & \textbf{0.906} & \textbf{0.052} & \textbf{0.758} & \textbf{0.762} & \textbf{0.008} & \textbf{0.840} & \textbf{0.840} & \textbf{0.049} \\ 

\hline\hline

\end{tabular}
\end{sc}
\end{small}
\end{center}
\end{table*}

\begin{table*}[t]
\caption{\textbf{[CheXpert - random label flipping]} Results for all 5 CheXpert tasks: Cardiomegaly (CA), Edema (ED), Consolidation (CO), Atelectasis (AT) and Pleural Effusion (PE). We see that randomly flipping the true labels of some fraction of a certain group has no clear trend and thus, while it may seem like a simple method to ensure fairness, it requires a lot of trial and error (\eg~ setting the right fraction to flip, which subgroup to flip \etc). There is also a decline in performance (AUC scores drop as compared to the baseline) for all of the label flip experiments.}
\label{tab:chexpert_random_flip}
\vskip 0.15in
\begin{center}
\begin{small}
\begin{sc}
\begin{tabular}{c|c|ccc|ccc|ccc}
\toprule
\multirow{2}{*}{Scheme} & \multirow{2}{*}{Task} & \multicolumn{3}{c|}{Train} & \multicolumn{3}{c|}{Validation} & \multicolumn{3}{c}{Test} \\
 &  & Male   & Female   & Gap    & Male     & Female     & Gap     & Male   & Female   & Gap  \\ 
\midrule
\midrule

\multirow{6}{*}{\begin{tabular}[c]{@{}c@{}}Baseline\end{tabular}} 
& CD & 0.988 & 0.990 & 0.002 & 0.826 & 0.818 & 0.007 & 0.766 & 0.739 & 0.027 \\
& ED & 0.953 & 0.952 & 0.000 & 0.780 & 0.777 & 0.002 & 0.885 & 0.862 & 0.024 \\ 
& CO & 0.996 & 0.996 & 0.000 & 0.681 & 0.687 & 0.006 & 0.896 & 0.808 & 0.088 \\ 
& AT & 0.879 & 0.883 & 0.004 & 0.637 & 0.631 & 0.007 & 0.637 & 0.570 & 0.068 \\ 
& PE & 0.936 & 0.942 & 0.006 & 0.841 & 0.854 & 0.013 & 0.895 & 0.925 & 0.030 \\
& \textbf{Avg} & \textbf{0.950} & \textbf{0.953} & \textbf{0.002} & \textbf{0.753} & \textbf{0.753} & \textbf{0.007} & \textbf{0.816} & \textbf{0.781} & \textbf{0.047} \\ 

\hline\hline

\multirow{6}{*}{\begin{tabular}[c]{@{}c@{}}Random Flip\\$\beta = 0.1$\\Male\\Flipped\end{tabular}}  
& CD & 0.980 & 0.943 & 0.037 & 0.814 & 0.794 & 0.020 & 0.743 & 0.790 & 0.047 \\ 
& ED & 0.930 & 0.882 & 0.048 & 0.756 & 0.750 & 0.006 & 0.892 & 0.774 & 0.118 \\ 
& CO & 0.967 & 0.897 & 0.070 & 0.656 & 0.636 & 0.020 & 0.600 & 0.667 & 0.067 \\ 
& AT & 0.853 & 0.797 & 0.056 & 0.622 & 0.594 & 0.028 & 0.668 & 0.612 & 0.056 \\ 
& PE & 0.925 & 0.914 & 0.011 & 0.841 & 0.850 & 0.009 & 0.832 & 0.916 & 0.084 \\
& \textbf{Avg} & \textbf{0.931} & \textbf{0.887} & \textbf{0.044} & \textbf{0.738} & \textbf{0.725} & \textbf{0.017} & \textbf{0.747} & \textbf{0.752} & \textbf{0.074} \\ 

\hline\hline

\multirow{6}{*}{\begin{tabular}[c]{@{}c@{}}Random Flip\\$\beta = 0.1$\\Female\\Flipped\end{tabular}}  
& CD & 0.902 & 0.965 & 0.063 & 0.738 & 0.778 & 0.039 & 0.711 & 0.646 & 0.065 \\ 
& ED & 0.890 & 0.919 & 0.029 & 0.735 & 0.744 & 0.009 & 0.908 & 0.827 & 0.081 \\ 
& CO & 0.836 & 0.908 & 0.072 & 0.580 & 0.643 & 0.064 & 0.685 & 0.668 & 0.017 \\ 
& AT & 0.777 & 0.825 & 0.048 & 0.610 & 0.606 & 0.004 & 0.687 & 0.758 & 0.071 \\ 
& PE & 0.877 & 0.909 & 0.032 & 0.807 & 0.833 & 0.026 & 0.776 & 0.927 & 0.150 \\
& \textbf{Avg} & \textbf{0.856} & \textbf{0.905} & \textbf{0.049} & \textbf{0.694} & \textbf{0.721} & \textbf{0.028} & \textbf{0.754} & \textbf{0.765} & \textbf{0.077} \\ 

\hline\hline

\multirow{6}{*}{\begin{tabular}[c]{@{}c@{}}Random Flip\\$\beta = 0.05$\\Male\\Flipped\end{tabular}}  
& CD & 0.989 & 0.961 & 0.028 & 0.790 & 0.762 & 0.028 & 0.716 & 0.634 & 0.082 \\ 
& ED & 0.950 & 0.921 & 0.030 & 0.757 & 0.744 & 0.013 & 0.889 & 0.742 & 0.147 \\ 
& CO & 0.971 & 0.941 & 0.030 & 0.684 & 0.681 & 0.003 & 0.876 & 0.758 & 0.118 \\ 
& AT & 0.854 & 0.827 & 0.027 & 0.652 & 0.632 & 0.020 & 0.689 & 0.675 & 0.014 \\ 
& PE & 0.917 & 0.910 & 0.007 & 0.829 & 0.841 & 0.013 & 0.842 & 0.906 & 0.063 \\
& \textbf{Avg} & \textbf{0.936} & \textbf{0.912} & \textbf{0.024} & \textbf{0.742} & \textbf{0.732} & \textbf{0.015} & \textbf{0.802} & \textbf{0.743} & \textbf{0.085} \\ 

\hline\hline

\multirow{6}{*}{\begin{tabular}[c]{@{}c@{}}Random Flip\\$\beta = 0.05$\\Female\\Flipped\end{tabular}}  
& CD & 0.956 & 0.979 & 0.024 & 0.802 & 0.811 & 0.009 & 0.729 & 0.726 & 0.003 \\ 
& ED & 0.906 & 0.924 & 0.017 & 0.730 & 0.737 & 0.007 & 0.869 & 0.733 & 0.137 \\ 
& CO & 0.897 & 0.952 & 0.055 & 0.622 & 0.635 & 0.013 & 0.659 & 0.624 & 0.035 \\ 
& AT & 0.760 & 0.784 & 0.024 & 0.592 & 0.592 & 0.001 & 0.676 & 0.639 & 0.037 \\ 
& PE & 0.872 & 0.883 & 0.011 & 0.787 & 0.799 & 0.012 & 0.795 & 0.765 & 0.030 \\
& \textbf{Avg} & \textbf{0.878} & \textbf{0.904} & \textbf{0.026} & \textbf{0.707} & \textbf{0.715} & \textbf{0.008} & \textbf{0.745} & \textbf{0.697} & \textbf{0.048} \\ 

\hline\hline

\multirow{6}{*}{\begin{tabular}[c]{@{}c@{}}Random Flip\\$\beta = 0.01$\\Male\\Flipped\end{tabular}}  
& CD & 0.984 & 0.981 & 0.003 & 0.786 & 0.776 & 0.010 & 0.782 & 0.683 & 0.099 \\ 
& ED & 0.949 & 0.941 & 0.008 & 0.765 & 0.756 & 0.009 & 0.856 & 0.779 & 0.076 \\ 
& CO & 0.974 & 0.972 & 0.002 & 0.647 & 0.643 & 0.003 & 0.792 & 0.794 & 0.002 \\ 
& AT & 0.821 & 0.814 & 0.007 & 0.621 & 0.608 & 0.013 & 0.770 & 0.770 & 0.000 \\ 
& PE & 0.915 & 0.917 & 0.002 & 0.817 & 0.820 & 0.003 & 0.834 & 0.916 & 0.082 \\
& \textbf{Avg} & \textbf{0.929} & \textbf{0.925} & \textbf{0.004} & \textbf{0.727} & \textbf{0.720} & \textbf{0.008} & \textbf{0.807} & \textbf{0.788} & \textbf{0.052} \\ 

\hline\hline

\multirow{6}{*}{\begin{tabular}[c]{@{}c@{}}Random Flip\\$\beta = 0.01$\\Female\\Flipped\end{tabular}}  
& CD & 0.981 & 0.988 & 0.007 & 0.793 & 0.799 & 0.006 & 0.720 & 0.661 & 0.060 \\ 
& ED & 0.925 & 0.928 & 0.003 & 0.720 & 0.713 & 0.007 & 0.840 & 0.815 & 0.025 \\ 
& CO & 0.962 & 0.972 & 0.010 & 0.670 & 0.692 & 0.022 & 0.733 & 0.718 & 0.015 \\ 
& AT & 0.742 & 0.752 & 0.010 & 0.547 & 0.552 & 0.005 & 0.492 & 0.585 & 0.093 \\ 
& PE & 0.806 & 0.827 & 0.021 & 0.714 & 0.729 & 0.015 & 0.714 & 0.848 & 0.134 \\
& \textbf{Avg} & \textbf{0.883} & \textbf{0.893} & \textbf{0.010} & \textbf{0.689} & \textbf{0.697} & \textbf{0.011} & \textbf{0.700} & \textbf{0.725} & \textbf{0.065} \\ 

\hline\hline

\end{tabular}
\end{sc}
\end{small}
\end{center}
\end{table*}

\end{document}